\pdfoutput=1

\documentclass[11pt]{article}

\usepackage[preprint]{acl}

\usepackage{times}
\usepackage{latexsym}

\usepackage[T1]{fontenc}

\usepackage[utf8]{inputenc}

\usepackage{microtype}

\usepackage{inconsolata}

\usepackage[pdftex]{graphicx}

\usepackage{subfigure}
\usepackage{booktabs}
\usepackage{hyperref}

\DeclareGraphicsExtensions{.pdf, .png}

\usepackage{algorithm}
\usepackage{algcompatible}

\usepackage{amsmath}
\usepackage{amssymb}
\usepackage{mathtools}
\usepackage{amsthm}
\usepackage[capitalize,noabbrev]{cleveref}

\theoremstyle{plain}

\theoremstyle{definition}

\theoremstyle{remark}

\usepackage[textsize=tiny]{todonotes}

\usepackage{xcolor}
\usepackage{colortbl}
\definecolor{hookersgreen}{rgb}{0.0, 0.44, 0.0}
\definecolor{indiagreen}{rgb}{0.07, 0.53, 0.03}
\definecolor{islamicgreen}{rgb}{0.0, 0.56, 0.0}
\definecolor{kellygreen}{rgb}{0.3, 0.73, 0.09}
\definecolor{alizarin}{rgb}{0.82, 0.1, 0.26}
\definecolor{lightgrey}{gray}{0.95}
\definecolor{middlegrey}{gray}{0.935}
\definecolor{lightcyan}{RGB}{204, 255, 255}
\definecolor{lightpurple}{RGB}{230, 225, 245}
\definecolor{lightpink}{RGB}{255,220,225} 
\definecolor{lightgreen}{RGB}{240,255,235}
\definecolor{lightyellow}{RGB}{255,255,230}

\usepackage{pifont}

\usepackage{makecell}
\usepackage{multirow}
\usepackage{arydshln}
\usepackage[shortlabels]{enumitem}

\title{MacRAG: Compress, Slice, and Scale-up \\for Multi-scale Adaptive Context RAG}

\author{
\begin{tabular}{c}
  Woosang Lim$^{1}$\thanks{Equal contribution.}\thanks{Corresponding authors: woosang.quasar@gmail.com, william@cs.ucsb.edu}\thanks{Work done while at POSCO HOLDINGS.} \,
  Zekun Li$^{2}$\footnotemark[1] \, 
  Gyuwan Kim$^{2}$\footnotemark[1] \, 
  Sungyoung Ji$^{1}$\footnotemark[1] 
  \\
  HyeonJung Kim$^{1}$ \,   
  Kyuri Choi$^{1}$ \, 
  Jin Hyuk Lim$^{1}$ \, 
  Kyungpyo Park$^{3}$ \, 
  William Yang Wang$^{2}$\footnotemark[2]
\end{tabular} \\
POSCO HOLDINGS$^{1}$ \, University of California, Santa Barbara$^{2}$ \, Google Cloud$^{3}$
}

\begin{document}

\maketitle

\begin{abstract}
Long-context large language models (LC LLMs) combined with retrieval-augmented generation (RAG) hold strong potential for complex multi-hop and large-document tasks. 
However, existing RAG systems often suffer from imprecise retrieval, incomplete context coverage under constrained windows, and fragmented information from suboptimal context construction. 
We introduce \textbf{Multi-scale Adaptive Context RAG (MacRAG)}, a hierarchical RAG framework that compresses and partitions documents into coarse-to-fine granularities, then adaptively merges relevant contexts through real-time chunk- and document-level expansions. 
By initiating with finest-level retrieval and progressively incorporating broader, higher-level context, MacRAG constructs effective query-specific long contexts, optimizing both precision and coverage. 
Evaluations on challenging LongBench expansions of HotpotQA, 2WikiMultihopQA, and Musique confirm MacRAG consistently surpasses baseline RAG pipelines in single- and multi-step generation using Llama-3.1-8B, Gemini-1.5-pro, and GPT-4o. 
Our results establish MacRAG as an efficient, scalable solution for real-world long-context, multi-hop reasoning.
Our code is available at \url{https://github.com/Leezekun/MacRAG}.

\end{abstract}

\section{Introduction}
\label{introduction}

\begin{figure*}[t]
\begin{center}
\includegraphics[width=\linewidth]{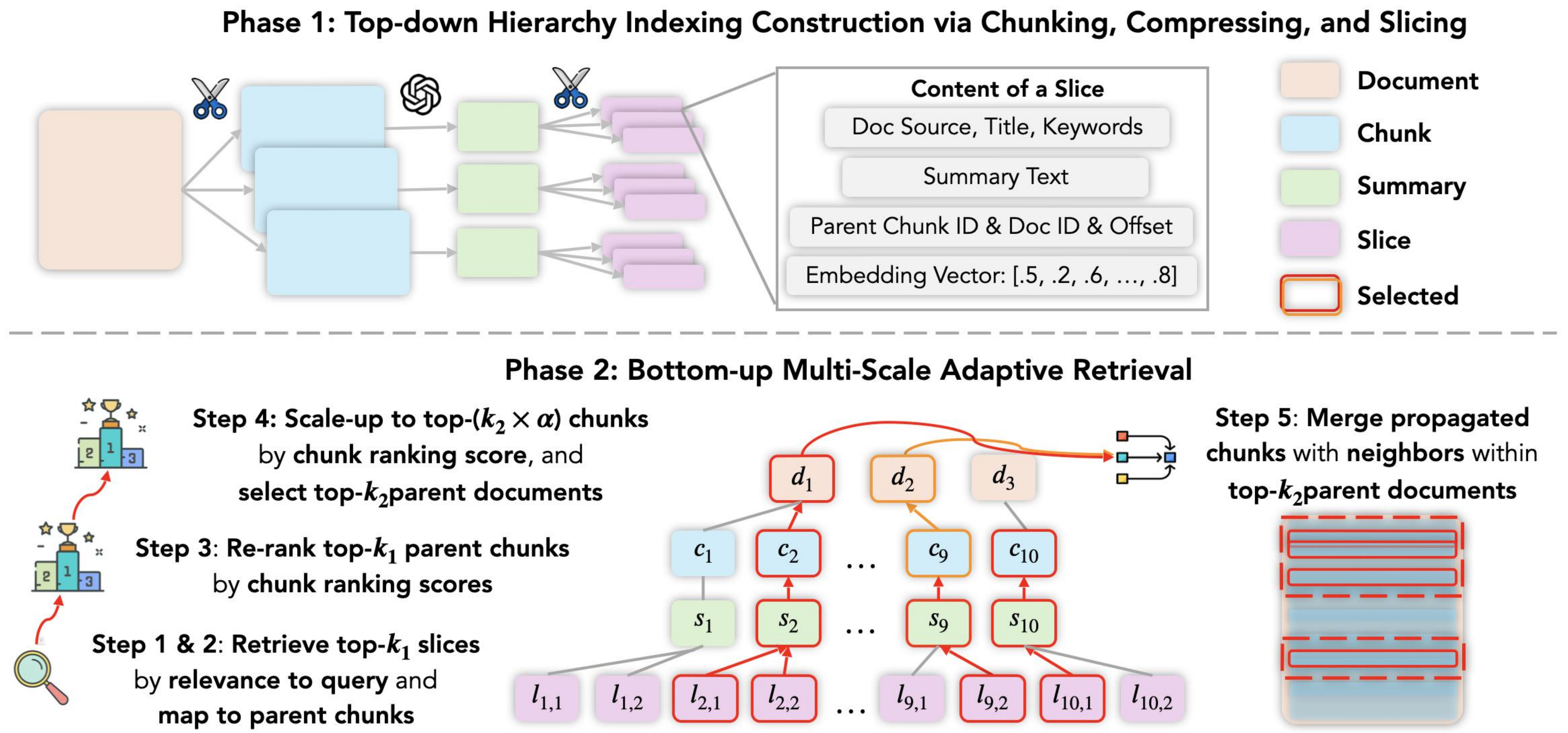}
\end{center}
\caption{An overview of the MacRAG framework, consisting of two main phases: (1) top-down hierarchical indexing (upper), and (2) bottom-up multi-scale adaptive retrieval on the constructed hierarchy of document-chunk-summary-slice (lower).
}
\label{fig_MacRAG_illustration}
\vspace{-2mm}
\end{figure*}

Large language models (LLMs) have significantly advanced complex reasoning, but they still suffer from factual gaps or hallucinations when relying only on internal parameters~\citep{zhao2024longrag}.  
Retrieval-Augmented Generation (RAG) addresses this by grounding LLMs in external evidence~\citep{guu2020retrieval, lewis2020retrieval}.  
Long-context (LC) LLMs such as GPT-4o~\citep{openai2024gpt4o}, Gemini 1.5~\citep{team2024gemini}, and Llama 3~\citep{dubey2024llama3} offer large input windows but remain limited.  
They often miss crucial mid-context information~\citep{liu2024lost, xu2024retrieval} and their performance degrades at extreme context lengths~\citep{leng2024long, yu2024defense}.

These issues in RAG systems give rise to three key trade-offs:  
(1) Context Length vs. Focus, where longer contexts improve recall but may obscure important details, while shorter ones enhance focus but risk omission of essential evidence;  
(2) Chunking and Indexing, where fine-grained chunks boost retrieval precision but harm coherence, while coarse chunks preserve structure but introduce redundancy; and  
(3) Coverage vs. Computation, where broader context improves reasoning but increases token cost and latency, especially in iterative or agentic pipelines~\citep{yue2024inference, asai2024selfrag}.  

To address these interconnected trade-offs systematically, we propose Multi-scale Adaptive Context RAG (MacRAG). 
MacRAG integrates top-down offline indexing with bottom-up query-time adaptive retrieval. Offline, documents are partitioned into overlapping chunks, their content compressed via abstractive summarization, and these summaries are further sliced for fine-grained indexing. 
At query time, MacRAG retrieves precise slices, then adaptively reconstructs the context by merging these into parent chunks, incorporating neighboring chunks, and performing document-level expansions. 
This constructs effective, query-specific, and length-bounded long contexts, optimizing the balance between precision, coverage, and efficiency.

MacRAG's unified approach combines structure-preserving indexing, adaptive multi-scale retrieval, and bounded context assembly, offering a distinct, robust solution for complex multi-hop reasoning. 
Extensive experiments on challenging LongBench~\citep{bai2023longbench} datasets show significant gains over strong baselines using Llama-3.1-8B, Gemini-1.5-pro, and GPT-4o.  
This paper thus introduces a novel multi-scale retrieval architecture, empirically validates its significant benefits for demanding QA tasks, and presents an efficient, modular framework for advanced RAG applications.
\section{Related Work}
\label{related_work}

Strategies for RAG with Long Context (LC) LLMs \citep{openai2024gpt4o, team2024gemini, dubey2024llama3} largely follow two directions. Post-retrieval context management aims to condense information after initial retrieval by employing methods such as abstractive summarization like RECOMP-Abst \citep{xu2024recomp} or extractive techniques including LLMLingua \citep{jiang2023llmlingua} and RECOMP-Extr \citep{xu2024recomp}. While these manage context size, they are bottlenecked by initial retrieval quality, risk information loss, and can be computationally intensive. MacRAG, by contrast, distinctly improves the retrieval phase itself, proactively constructing a high-quality, bounded context.

Orthogonally, hierarchical retrieval approaches organize information during retrieval and ranking for broader coverage. Systems like GraphRAG \citep{edge2024local} and HippoRAG \citep{gutierrez2024hipporag} use symbolic graphs or sentence-level indexing, which can improve recall but often increase the overhead. RAPTOR \citep{sarthiraptor} recursively summarizes clustered text chunks, potentially missing non-semantic relations, while SIRERAG \citep{zhang2025sirerag} further integrates relational connectivity at higher computational cost. LongRAG \citep{zhao2024longrag} utilizes entire parent documents of top-ranked chunks alongside a multi-step generation scheme, a strategy that can be inefficient with very long document contexts.

Based on \citet{zhang2025sirerag}'s comparative analysis (Table 4 therein), which shows that GraphRAG underperforms on multi-hop QA while RAPTOR, HippoRAG, and SIRERAG attain similar F1-scores, we focus our evaluations on RAPTOR and LongRAG. Although RAPTOR’s semantic clustering may fragment knowledge and LongRAG’s full-document usage can be costly, both were selected for their promising trade-offs between strong performance and relatively lower overhead compared to other competitive methods, such as the noted underperformance of GraphRAG or the higher operational costs of SIRERAG.

In contrast, MacRAG preserves original document structure through its offline hierarchical indexing and adaptively merges relevant and related contexts via its multi-scale retrieval and ranking at query time. This approach circumvents costly, repeated clustering and the need for explicit symbolic graphs, offering flexible assembly of multi-hop contexts with minimal overhead, and is designed with extendibility towards graph-based enhancements.

\section{Multi-scale Adaptive Context RAG}
\label{method}

\begin{algorithm}[t]
\footnotesize
\caption{\textbf{MacRAG}: Multi-scale Adaptive Context RAG}
\label{alg_macrag_compact}
\begin{algorithmic}
\REQUIRE{Query $q$, document corpus $\mathcal{D}$, $k_1$ (for slice retrieval), $k_2$ (for final merged chunks), the \# of hops $h$, up-scaling factor $\alpha$}
\ENSURE{Top-$k_{2}$ merged chunks $\mathcal{C}_{\mathrm{final}}$, generated answer}
\vspace{1mm}
\STATE \textbf{{Phase 1: Top-down Hierarchical Indexing}}
\STATE For each document $d \in \mathcal{D}$:
\STATE \; \textbf{(1.1)} Split $d$ into a set of overlapping \emph{chunks} $\mathcal{C}_d$.
\STATE \; \textbf{(1.2)} \emph{Compress} each chunk $c_i$ into a shorter summary $s_i$ via summarization.
\STATE \; \textbf{(1.3)} Split each compressed summary $s_i$ into overlapping \emph{slices} $\mathcal{L}_i$.
\STATE \; \textbf{(1.4)} Encode each slice and store its embedding \& metadata (doc ID, chunk ID, offset, etc) in the database.
\vspace{1mm}
\STATE \textbf{{Phase 2: Bottom-up Multi-scale Adaptive Retrieval and Ranking}}
\STATE \; \textbf{(2.1)} Retrieve the top-$k_{1}$ slices based on similarity to $q$.
\STATE \; \textbf{(2.2)} Map retrieved slices to their parent chunks to obtain $\mathcal{C}_{k_{1}}'$.
\STATE \; \textbf{(2.3)} \emph{Rerank} chunks in $\mathcal{C}_{k_{1}}'$ to refine their scores $r'_{q,c_i}$.
\STATE \; \textbf{(2.4)} \emph{Scale-up} by selecting the top $(k_{2} \times \alpha)$ chunks and ranking their source documents by aggregated scores to choose $k_{2}$ distinct documents.
\STATE \; \textbf{(2.5)} \emph{Merge} the top-ranked chunks' $h$-hop neighbors to form the final top-$k_2$ chunks, $\mathcal{C}_{\mathrm{final}}$.
\vspace{1mm}
\STATE \textbf{{Phase 3: Response Generation}}  
\STATE Provide $\mathcal{C}_{\mathrm{final}}$ as context to an LLM (or any downstream model) to generate the final answer. 
\end{algorithmic}
\end{algorithm}

We propose \textbf{Multi-scale Adaptive Context RAG (MacRAG)}, a hierarchical multi-scale adaptive retrieval system with three sequential components. 
First, \textbf{top-down hierarchical indexing} compresses documents, partitions them into chunks, and slices them to build a hierarchical index from coarse-grained documents to fine-grained slices with multi-level indices. 
Second, \textbf{bottom-up multi-scale adaptive retrieval} starts from the finest granularity to ensure precision, and progressively expands to broader contexts by merging relevant information to construct \textbf{query-specific long contexts dynamically} in real time.  
Third, the \textbf{response generation} leverages these carefully assembled contexts to produce the answer.

Our main contribution lies in the retrieval, ranking, and adaptive input context construction pipeline (phases 1 and 2), which serves as a modular foundation for integrating any response generation method in phase 3.  
Figure~\ref{fig_MacRAG_illustration} provides an overview of MacRAG, and Algorithm~\ref{alg_macrag_compact} outlines the methods along with equations in subsequent sections.

\subsection{Document Hierarchical Indexing}
\label{subsec_offline}
Given a document corpus $\mathcal{D} = \{d_1, \dots, d_N\}$, MacRAG performs offline processing to construct a hierarchy through chunking (\S\ref{sec:chunking}), compression (\S\ref{sec:compression}), and slicing (\S\ref{sec:slicing}), while building a \emph{slice-level} vector DB at the finest granularity to empower precise retrieval. 
This strategy aims to reduce redundancy and sharpen semantic focus, thereby enhancing retrieval precision and minimizing noise during the subsequent multi-scale expansion phase.

\subsubsection{Chunking}
\label{sec:chunking}
Each document $d \in \mathcal{D}$ is split into partially overlapping chunks, $\mathcal{C}_d =\{c_1,c_2,\dots,c_m \}$,
where each chunk $c_i$ spans approximately 200 to 500 tokens, with a fixed overlap of 10–200 tokens between consecutive chunks. This controlled overlap preserves semantic continuity across chunk boundaries. Although the chunking is token-based by default, it can alternatively be configured at the character level. 
We applied the similar range of chunk sizes used by LongRAG ~\citep{zhao2024longrag}, more details in Appendix \ref{sec:appendix_efficiency_MacRAG}.

\subsubsection{Compression}
\label{sec:compression}
Each chunk $c_i$ is transformed into a more compressed summary $s_i$, yielding a set of summaries $\mathcal{S}_d = \{s_1,\dots,s_m\}$, where 
$s_i = \mathrm{Compress}(c_i)$. 
The $\mathrm{Compress}(\cdot)$ represents a summarization model.
It can be either extractive (selecting salient sentences) or abstractive (using a generative model to generate a summary). By default, we use an LLM to generate the abstractive summarization. This step effectively reduces redundancy while retaining core factual information.

\subsubsection{Slicing}
\label{sec:slicing}
To facilitate finer-grained retrieval, each summary $s_i$ is further split into overlapping slices:
$\mathcal{L}_i = \bigl\{\,\ell_{i,1},\ell_{i,2},\dots,\ell_{i,j},\dots\bigr\}$,
where each slice $\ell_{i,j}$ typically spans $50\text{-}200$ tokens with partial overlap as in the standard overlap chunking, and our parameter details are in Appendix \ref{sec:appendix_efficiency_MacRAG}.
Each slice is then embedded via $\mathbf{e}_{i,j} = \mathrm{Embed}(\ell_{i,j})$,
where $\mathrm{Embed}(\cdot)$ represents a text encoder (\textit{e.g.}, \texttt{multilingual-e5-large}). The slice-level embeddings with metadata (document ID, chunk ID, offsets) are indexed in the vector DB. 
MacRAG can optionally index chunk and summary embeddings for multi-view retrieval.

\subsection{Multi-scale Adaptive Retrieval and Ranking}
\label{subsec_online}
Given an input query $q$, MacRAG implements a multi-scale adaptive retrieval pipeline over the constructed hierarchy through five sequential steps, ultimately selecting the top-$k_2$ merged chunks with contextually relevant neighbor expansions for enhanced both precision and recall.

\subsubsection{Initial Slice-level Retrieval}
With slices serving as the finest granular units in our hierarchical system and vector DB, we first compute the query-slice relevance scores for each slice $\ell_{i,j}$ in the vector DB: $r_{q,\ell_{i,j}} = \mathrm{Rel}\!\bigl(q,\,\ell_{i,j}\bigr)$,
where $\mathrm{Rel}(\cdot,\cdot)$ can be implemented with relevance model, \textit{e.g.}, cosine similarity and $L_2$ distance in dense retrieval, sparse retrieval, hybrid retrieval, and so on.
We then retrieve the top-$k_1$ most relevant slices.
This fine-grained retrieval ensures high precision by identifying slices with strong semantic alignment to the query $q$.

\subsubsection{Parent Chunk Mapping}
Each retrieved slice corresponds to a parent chunk $c_i$. To prevent redundancy when multiple retrieved slices originate from the same chunk, we perform a unique mapping: $\mathcal{C}_{k_1}^{'} = \mathrm{unique}\bigl({c_i\mid \ell_{i,\cdot}\in \mathrm{top}\text{-}k_1(r_{q, \ell})}\bigr)$,
where we identify and consolidate the parent chunks of all retrieved slices and obtain a set of selected chunks used for the next steps.

\subsubsection{Chunk-level Re-Ranking}
For each identified unique relevant chunk from the mapping as $c_i\in\mathcal{C}_{k_1}^{'}$ to the query, we compute a chunk-level ranking score for the query: $r_{q,c_i} = \mathrm{ReRank}\bigl(q,\,c_i\bigr)$,
utilizing a cross-encoder (\textit{e.g.}, \texttt{marco-miniLM}) or advanced ranking model that evaluates the complete chunk content, mitigating potential ``information fragmentation" inherent in slice-level retrieval. The chunks in $\mathcal{C}_{k_1}^{'}$ are then reordered according to their ranking scores to the query $q$.

\subsubsection{Scaling-Up \& Document-level Ranking}
This step aims to select optimal content segments for constructing $k_2$ precise, bounded long contexts for the final LLM generation, ensuring both relevance and sufficient coverage. 
It incorporates top-ranked chunks and their context neighbors in the same document to mitigate information fragmentation and support multi-hop reasoning.

\textbf{Scaled Top Chunk Selection.~} 
Recognizing that initial re-ranking scores $r_{q,c_i}$ may not be perfect, we consider a broader set of borderline candidates rather than strictly picking the top-$k_2$ chunks. 
MacRAG employs a scaling factor $\alpha \geq 1$ (where $(k_{2} \times \alpha) \leq |\mathcal{C}^{'}_{k_1}|$) to select the top-($k_{2} \times \alpha$) chunks from $\mathcal{C}_{k_1}^{'}$: $\mathcal{C}^{'}_{q, k_2} = \mathrm{TopK}_{(k_{2} \times \alpha)}\!\bigl(r_{q,c_i}\bigr) \text{~for~} c_i \in \mathcal{C}^{'}_{q, k_1}$. 
$\alpha = 1$ does not scale-up the number of borderline candidates, but 
moderate higher $\alpha$ (\textit{e.g.}, 3 or 4) can provide higher probabilities to include borderline but crucial chunks to improve coverage while maintaining a certain level of quality for candidates. 
To balance among precision, recall, and context lengths, $\alpha \in \{ 2, 3, 4\}$ would be promising where $(k_{2} \times \alpha) \leq |\mathcal{C}^{'}_{k_1}|$, and we will also demonstrate that selection $\alpha$ is not sensitive, but robust with handling trade-off problems well in Section \ref{experiments}.
This approach can also potentially facilitate bridging multi-hop queries.

\textbf{Document Selection.~} 
From the scaled set of $(k_{2} \times \alpha)$ chunks in $\mathcal{C}^{'}_{q, k_2}$, we identify their unique source documents. To select the top-$k_2$ documents, we compute a document-level rank score $r_{q,d}$ for each candidate document $d$. 
This score $r_{q,d}$ leverages the chunk-level scores $r_{q,c_i}$ of document's all constituent chunks $c_i$ that are also present in $\mathcal{C}^{'}_{q, k_2}$: $r_{q,d} = \mathrm{ReRank_{Doc}}\Bigl( \{ c_i \in d \text{~and~} c_i \in \mathcal{C}^{'}_{q, k_2} \}\Bigr)$. 
Document ranking functions can include \texttt{mean}, \texttt{max}, \texttt{sum} of chunk scores, or an LLM-based ranker (with $\max(r_{q,c_i})$ being a suitable base if using cross-encoder scores, which can be either positive or negative). 
Based on $r_{q,d}$, we form $\mathcal{D}_{q, k_2}$, a set of top-$k_2$ \emph{distinct} documents. 
Ensuring document diversity, especially with appropriate $\alpha$ values (\textit{e.g.}, 3 or 4 depending on the overlapped ratio of original chunks), mitigates the risk of concentrating on too few sources and is critical when evidence is scattered across multiple documents.

\subsubsection{Neighbor Chunk Propagation \& Merge for Input Context Construction}
Within each of the top-$k_2$ \emph{distinct} documents, we focus on chunks $c_i$ both in $\mathcal{D}_{q, k_2}$ and $\mathcal{C}^{'}_{q, k_2}$, as these represent the most relevant content segments. 
For such a chunk $c_i$, we define $\mathcal{N}_{h}(c_i)$ which consists of $c_i$'s neighbors within $h$-hops and $c_i$ itself, and process extension of $c_i$ to get $\mathcal{N}(c_i)$ by using its associated indices, $\mathcal{N}_{h}(c_i) = \;\bigl\{\,c_{i-h},\dots,c_i,\dots,c_{i+h}\bigr\}$.
We then merge $\mathcal{N}_h(c_i)$ within the same ranked document to form the final merged chunk for adaptive long context to the query $q$: $c^{merge}_{q,d} = \mathrm{Merge}\bigl(\mathcal{N}_h(c_i)\bigr)$, where $d \in \mathcal{D}_{q, k_2}$, $c_i \in d$, and $c_i \in \mathcal{C}^{'}_{q, k_2}$.
From the top-$k_2$ documents in $\mathcal{D}_{q, k_2}$, we can select top-$k_2$ merged chunks  $c^{merge}_{q,d}$ by considering their order along with the top document order with $r_{q,d}$.
Then, finally, we have top-$k_2$ merged chunks as $\mathcal{C}_{\mathrm{final}} = \mathrm{TopK}_{k_2}\!\Bigl(c^{merge}_{q,d}\Bigr)$.
The final merged chunks $\mathcal{C}_{\text{final}}$ provide diverse views across the top-$k_2$ documents while optimizing context length \citep{zhao2024longrag}, combining the most relevant content with minimal redundancy. 

\subsection{Single/Multi-step Post Retrieval Generation}  %
\label{subsec_generation_modes}
Our retrieval framework is modular, enabling integration with various post-retrieval generation approaches. These include both vanilla single-step generation directly using the retrieved top-$k_2$ chunks in a single pass and iterative multi-step generation as in LongRAG~\citep{zhao2024longrag}, which introduces a CoT-guided filter (which checks chunk-level relevance) and an LLM-augmented extractor (which locates global context in original passages). Following the comprehensive exploration of single and multi-step generation variants in LongRAG, we evaluate MacRAG's performance when combined with different post-retrieval generation methods.
Specifically, we investigate seven generation methods, as detailed in Table~\ref{tbl_generation_modes}. Five of these methods (\texttt{R\&B}, \texttt{R\&L}, \texttt{Full\_Ext}, \texttt{Fil}, and \texttt{Full\_E\&F}) are adopted directly from LongRAG. Among these, \texttt{R\&L}, \texttt{Full\_Ext}, and \texttt{Full\_E\&F} process the entire set of top-$k_2$ documents, which can introduce computational overhead when handling very long texts. To address this limitation while maintaining an optimal balance between precision and recall, we introduce two new variants: \texttt{R\&B\_Ext} and \texttt{R\&B\_E\&F}, to reduce this overhead while maintaining precision-recall balance.

\subsection{Implementation Complexity and System Efficiency}
MacRAG enhances simple overlapping-chunk indexing by first chunking each document with partial overlaps, then compressing each chunk through summarization, and finally slicing it. 
These processes are performed once during offline preparation, eliminating query-time overhead.
Moreover, this document-level independence simplifies updates and streamlines database operations, avoiding the complexity associated with hierarchical multi-clustering \cite{sarthiraptor, zhang2025sirerag}.

At inference time, MacRAG constructs a real-time \textit{effective long context} by mapping slices back to their parent chunks and documents using lightweight index-based lookups rather than costly tree traversals. This design leverages metadata dictionaries (document ID, chunk ID, token offsets) to enable modular neighbor merging with minimal overhead. A bounded context size is enforced through parameters \((k_{2} \times \alpha)\) and chunk dimensions, ensuring scalability to enterprise-level corpora without degrading performance. 
This controlled scalability and real-time context assembly underscore MacRAG's value for massive long-document collections, ``almost infinite context" scenarios in advanced personalization, and as a robust foundation for iterative RAG or agentic systems.
\begin{table*}[t!]
\centering
\fontsize{8}{9.5}\selectfont
\begin{tabular}{l | c c c|c}
\hline
\textbf{Model} & \textbf{HotpotQA} & \textbf{2WikimultihopQA} & \textbf{Musique} & \textbf{Average} \\
\hline

\multicolumn{5}{ c }{\textbf{Long Context (LC) LLM without RAG}} \\
Gemini-1.5-pro \cite{team2024gemini} & 36.79 & 38.31 & 20.09 & 31.73 \\
GPT-4o \cite{openai2024gpt4o} & 46.76 & 40.62 & 30.76 & 39.38 \\
\hline

\multicolumn{5}{ c }{\textbf{Agentic RAG Method}} \\
CRAG (\textit{GPT-3.5-Turbo}) \cite{yan2024corrective} & 52.04 & 41.13 & 25.34 & 39.50 \\
Self-RAG (\textit{GPT-3.5-Turbo}) \cite{asai2024selfrag} & 50.51 & 46.75 & 24.62 & 40.63 \\
\hline

\multicolumn{5}{ c }{\textbf{RAG with Reranking (\texttt{R\&B}, Base Version)}} \\
Llama3-8B-8k \cite{dubey2024llama} & 48.25 & 43.47 & 19.66 & 37.13 \\
GPT-3.5-Turbo & 52.31 & 43.44 & 25.22 & 40.32 \\

\rowcolor{lightgrey}
Llama-3.1-8B-instruct (abbr. Llama-3.1-8B) & {52.50} & {46.33} & {26.70} & {41.84} \\
\rowcolor{lightgrey}
{RAPTOR \cite{sarthiraptor} (Llama-3.1-8B)}  & {52.30} ({-0.20$\downarrow$}) & {43.61} ({-2.72$\downarrow$})  & {23.79} ({-2.91$\downarrow$})  & {39.90} ({-1.94$\downarrow$}, {-4.64\%$\downarrow$})  \\  

\rowcolor{lightgrey}
\textbf{MacRAG (Ours)} (Llama-3.1-8B)  & \textbf{57.39} (\textbf{+4.89$\uparrow$}) & \textbf{44.87} (\textbf{-1.46$\downarrow$}) & \textbf{30.38} (\textbf{+3.68$\uparrow$}) & \textbf{44.21} (\textbf{+2.37$\uparrow$}, \textbf{+5.66\%$\uparrow$}) \\

\hline
\multicolumn{5}{ c }{\textbf{LongRAG \cite{zhao2024longrag} (Multi-step  Generation, \texttt{Full\_E\&F} in Table \ref{tbl_generation_modes})}} \\
LongRAG (GPT-3.5-Turbo) & 56.17 & 51.37 & 32.83 & 46.79 \\

\rowcolor{lightgrey}
LongRAG (Llama-3.1-8B) & {58.49} & {55.90} & {32.73} & {49.04} \\
\rowcolor{lightgrey}
\textbf{MacRAG (Ours)} (Llama-3.1-8B)  & \textbf{61.10} (\textbf{+2.61$\uparrow$}) & \textbf{57.74} (\textbf{+1.84$\uparrow$}) & \textbf{34.43} (\textbf{+1.70$\uparrow$}) & \textbf{51.09} (\textbf{+2.05$\uparrow$}, \textbf{+4.18\%$\uparrow$}) \\

\rowcolor{lightyellow}
LongRAG (Gemini-1.5-pro) & {61.95}  & {58.17}  & {34.04}  & {51.39}  \\
\rowcolor{lightyellow}
\textbf{MacRAG (Ours)} (Gemini-1.5-pro) & \textbf{64.39} (\textbf{+2.44$\uparrow$}) & \textbf{64.75} (\textbf{+6.58$\uparrow$}) & \textbf{43.34} (\textbf{+9.30$\uparrow$}) & \textbf{57.49} (\textbf{+6.10$\uparrow$}, \textbf{+11.87\%}$\uparrow$) \\

\rowcolor{lightcyan}
LongRAG  (GPT-4o) & {66.20}  & {65.89}  & {43.83}  & {58.64}  \\
\rowcolor{lightcyan}
\textbf{MacRAG (Ours)}  (GPT-4o) & \textbf{68.52} (\textbf{+2.32$\uparrow$}) & \textbf{73.19} (\textbf{+7.30$\uparrow$}) & \textbf{50.09} (\textbf{+6.26$\uparrow$}) & \textbf{63.93} (\textbf{+5.29$\uparrow$}, \textbf{+9.02\%$\uparrow$}) \\

\hline
\end{tabular}
\caption{
\label{tbl_experimental_results_MacRAG_vs_LongRAG_and_others}
Experimental results (F1-score) comparing \textbf{RAPTOR}, \textbf{LongRAG} and \textbf{MacRAG} on HotpotQA, 2WikimultihopQA, and MuSiQue datasets from LongBench~\cite{bai2023longbench}. Gains are displayed in parentheses with absolute number and its relative percentage. 
}
\end{table*}

\begin{table*}[ht!]
\centering
\scriptsize
\renewcommand{\arraystretch}{1.1}
\setlength{\tabcolsep}{5pt}
\begin{tabular}{c|c|c|c|c|c}
\hline
\multirow{2}{*}{\textbf{Dataset}} 
 & \textbf{Avg. Performance on} 
 & \textbf{LongRAG} 
 & \textbf{MacRAG} 
 & \textbf{LongRAG} 
 & \textbf{MacRAG} \\
 & \textbf{Seven Gen. Settings} 
 & \textbf{(Gemini-1.5-pro)} 
 & \textbf{(Gemini-1.5-pro)} 
 & \textbf{(GPT-4o)} 
 & \textbf{(GPT-4o)} \\
\hline
\multirow{4}{*}{HotpotQA} 
 & \textbf{Exact Match} 
   & 46.48 
   & \textbf{48.36} (\textbf{+1.88}$\uparrow$, \textbf{+4.04\%}$\uparrow$) 
   & 51.85 
   & \textbf{53.60} (\textbf{+1.75}$\uparrow$, \textbf{+3.37\%}$\uparrow$) \\
 & \textbf{F1-score} 
   & 61.45 
   & \textbf{63.97} (\textbf{+2.52}$\uparrow$, \textbf{+4.10\%}$\uparrow$) 
   & 66.17 
   & \textbf{67.51} (\textbf{+1.34}$\uparrow$, \textbf{+2.03\%}$\uparrow$) \\
 & \textbf{Precision} 
   & 65.93 
   & \textbf{67.98} (\textbf{+2.05}$\uparrow$, \textbf{+3.11\%}$\uparrow$) 
   & 68.66 
   & \textbf{70.10} (\textbf{+1.44}$\uparrow$, \textbf{+2.10\%}$\uparrow$) \\
 & \textbf{Recall} 
   & 61.59 
   & \textbf{64.47} (\textbf{+2.88}$\uparrow$, \textbf{+4.68\%}$\uparrow$) 
   & 68.10 
   & \textbf{69.36} (\textbf{+1.26}$\uparrow$, \textbf{+1.85\%}$\uparrow$) \\
\hline
\multirow{4}{*}{2WikimultihopQA} 
 & \textbf{Exact Match} 
   & 49.93 
   & \textbf{53.69} (\textbf{+3.76}$\uparrow$, \textbf{+7.53\%}$\uparrow$) 
   & 52.90 
   & \textbf{57.60} (\textbf{+4.69}$\uparrow$, \textbf{+8.87\%}$\uparrow$) \\
 & \textbf{F1-score} 
   & 57.73 
   & \textbf{61.97} (\textbf{+4.24}$\uparrow$, \textbf{+7.34\%}$\uparrow$) 
   & 62.69 
   & \textbf{67.45} (\textbf{+4.76}$\uparrow$, \textbf{+7.59\%}$\uparrow$) \\
 & \textbf{Precision} 
   & 57.97 
   & \textbf{61.69} (\textbf{+3.72}$\uparrow$, \textbf{+6.41\%}$\uparrow$) 
   & 62.22 
   & \textbf{66.48} (\textbf{+4.26}$\uparrow$, \textbf{+6.85\%}$\uparrow$) \\
 & \textbf{Recall} 
   & 60.17 
   & \textbf{65.10} (\textbf{+4.93}$\uparrow$, \textbf{+8.20\%}$\uparrow$) 
   & 66.47 
   & \textbf{72.00} (\textbf{+5.53}$\uparrow$, \textbf{+8.32\%}$\uparrow$) \\
\hline
\multirow{4}{*}{Musique} 
 & \textbf{Exact Match} 
   & 26.05 
   & \textbf{33.75} (\textbf{+7.70}$\uparrow$, \textbf{+29.57\%}$\uparrow$) 
   & 31.30 
   & \textbf{36.48} (\textbf{+5.18}$\uparrow$, \textbf{+16.55\%}$\uparrow$) \\
 & \textbf{F1-score} 
   & 33.92 
   & \textbf{42.84} (\textbf{+8.93}$\uparrow$, \textbf{+26.34\%}$\uparrow$) 
   & 41.43 
   & \textbf{47.91} (\textbf{+6.48}$\uparrow$, \textbf{+15.64\%}$\uparrow$) \\
 & \textbf{Precision} 
   & 34.69 
   & \textbf{43.29} (\textbf{+8.60}$\uparrow$, \textbf{+24.78\%}$\uparrow$) 
   & 40.80 
   & \textbf{47.07} (\textbf{+6.27}$\uparrow$, \textbf{+15.37\%}$\uparrow$) \\
 & \textbf{Recall} 
   & 35.69 
   & \textbf{45.16} (\textbf{+9.46}$\uparrow$, \textbf{+26.50\%}$\uparrow$) 
   & 45.06 
   & \textbf{51.67} (\textbf{+6.61}$\uparrow$, \textbf{+14.67\%}$\uparrow$) \\
\hline
\end{tabular}
\caption{
\label{tbl_average_performances_on_seven_generations}
Consolidated results for four metrics across three datasets, comparing LongRAG and MacRAG using the Gemini-1.5-pro and GPT-4o models. The results are averaged over seven generation settings in Table \ref{tbl_generation_modes}. 
}
\end{table*}

\begin{figure*}[t]
\centering
\includegraphics[width=0.32\textwidth]{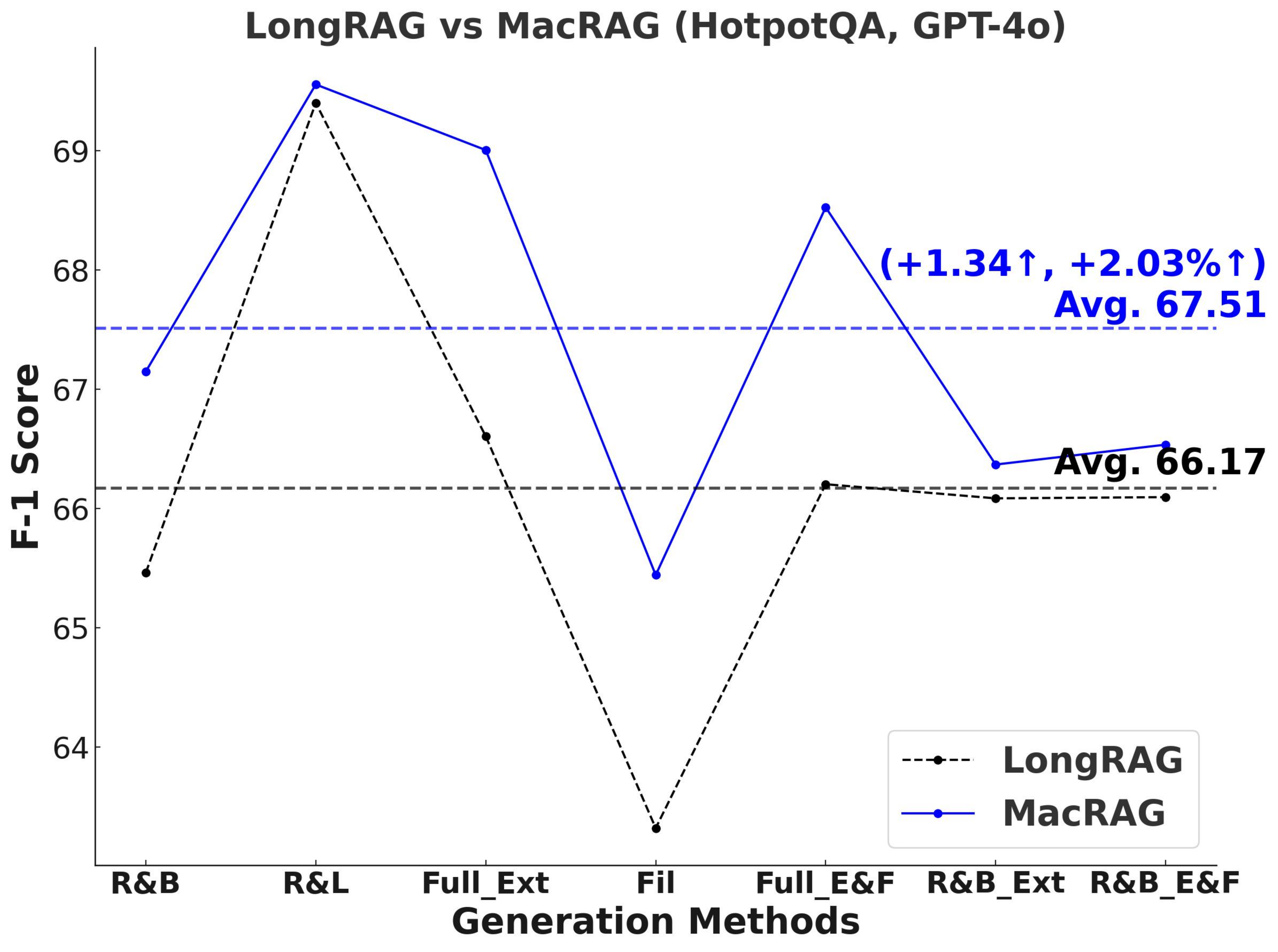}
\includegraphics[width=0.32\textwidth]{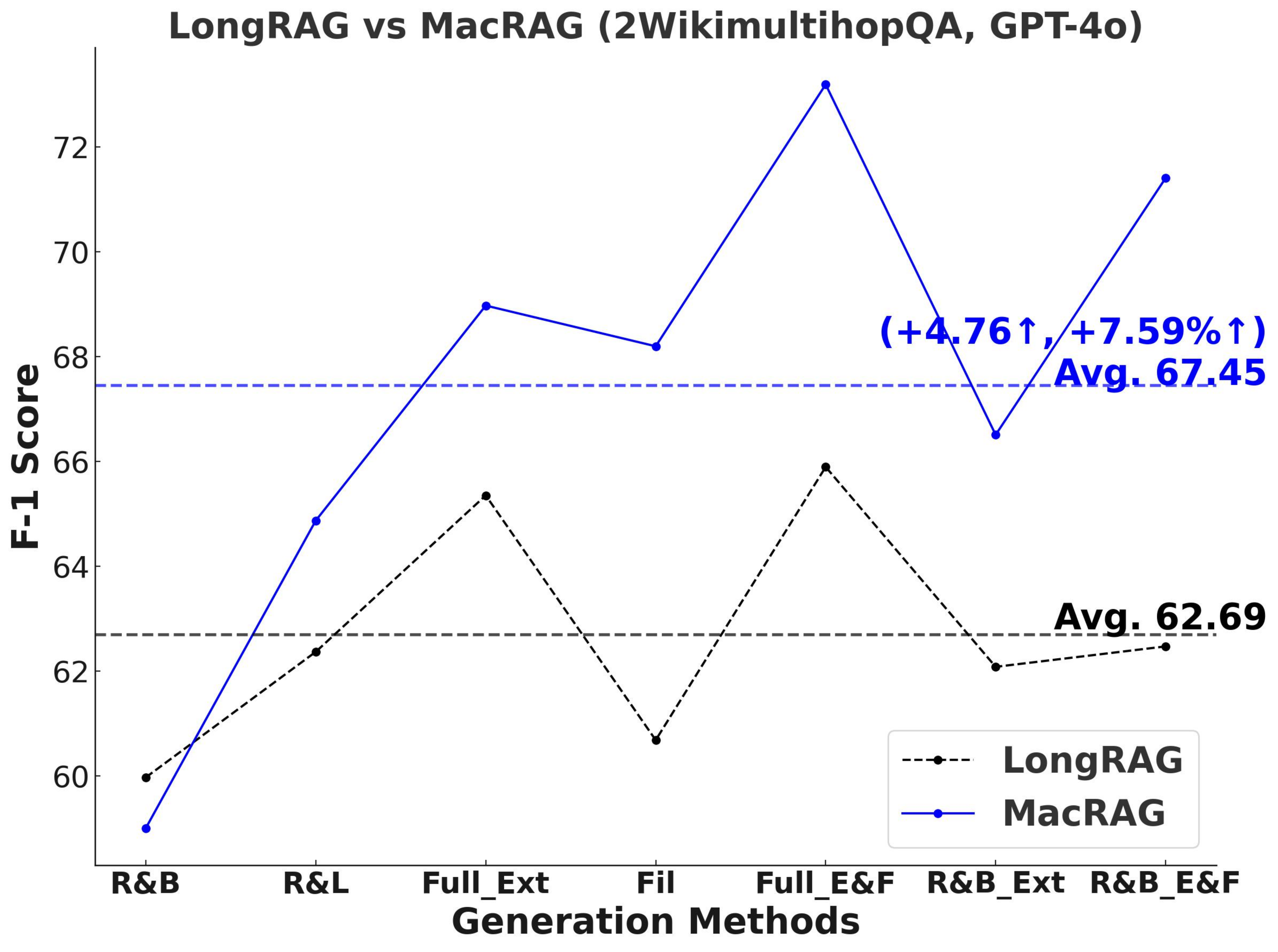}
\includegraphics[width=0.32\textwidth]{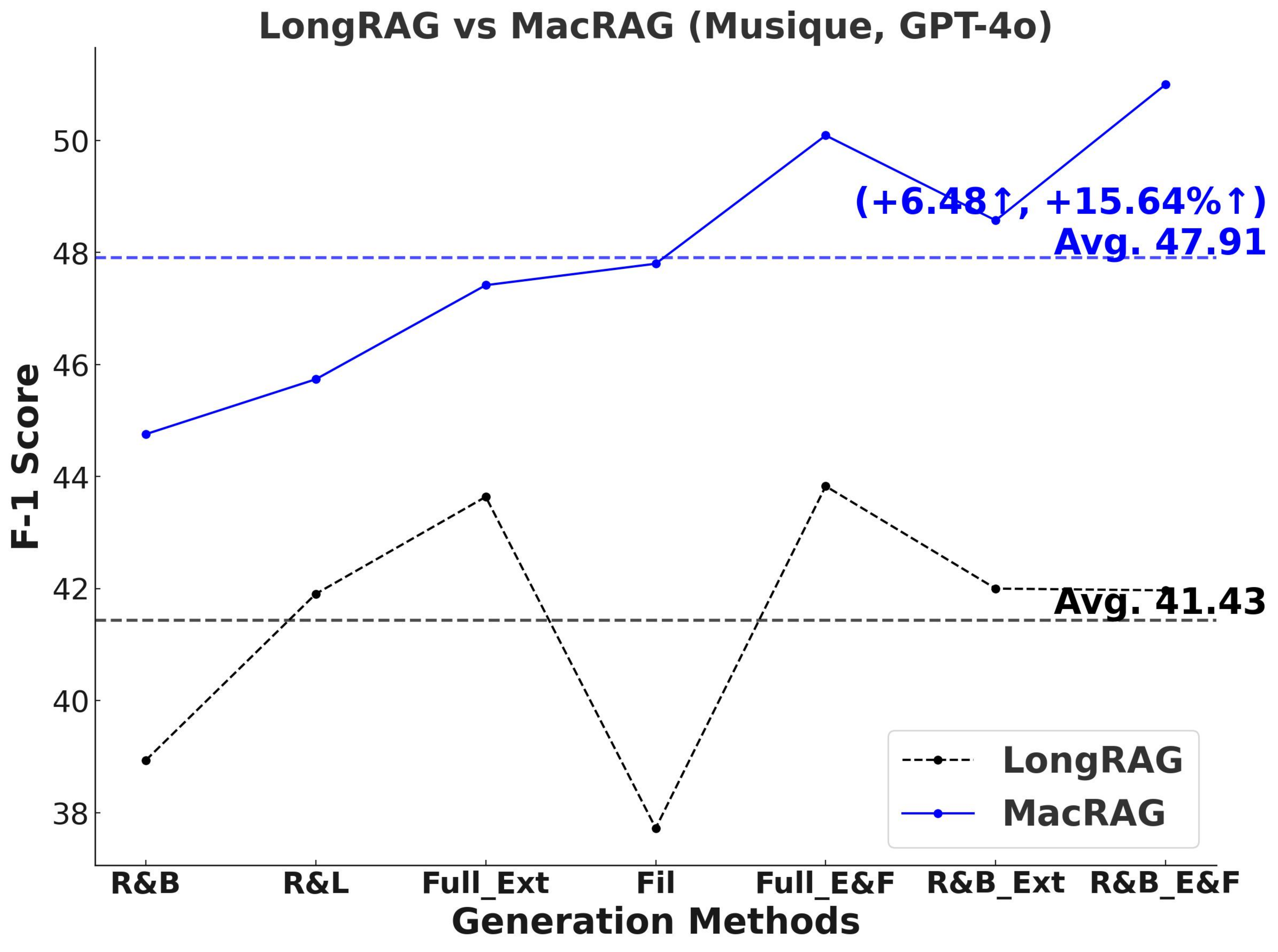}

\caption{Performance comparison of GPT-4o using LongRAG and MacRAG across the seven settings in Table~\ref{tbl_generation_modes}, showing F1-scores for three datasets (HotpotQA, 2WikiMultihopQA, and MuSiQue). Complete results for all metrics and LLMs are provided in the Appendix.}
\label{fig_seven_generation_method_all_datasets_for_Gemini_GPT_f1_score}
\end{figure*}

\section{Experiments}
\label{experiments}


\subsection{Experimental Setup}
We evaluate MacRAG on three challenging multi-hop question answering benchmarks from LongBench \citep{bai2023longbench}: HotpotQA, 2WikiMultihopQA, and Musique. These datasets, with their tangential passages and obscured connecting information, specifically test a system's ability to handle the "Lost in the Middle" phenomenon and perform robust multi-hop reasoning—precisely the challenges MacRAG addresses.
Section~\ref{sec:benchmark_datasets} in the Appendix provides a detailed explanation about these datasets.

\paragraph{Models and Evaluations.}
Our evaluations employ three prominent LC LLMs: GPT-4o \citep{openai2024gpt4o}, Gemini-1.5-pro \citep{team2024gemini}, and the open-source Llama-3.1-8B-instruct \citep{dubey2024llama3}. Performance is measured using Exact Match (EM), F1-score, Precision, and Recall. We primarily compare MacRAG against strong baselines RAPTOR \citep{sarthiraptor} and LongRAG \citep{zhao2024longrag}, selected for their representative state-of-the-art performance and efficiency balance in hierarchical RAG (as discussed in Section \ref{related_work}). 
To ensure rigorous comparative analysis across all RAG methods (MacRAG, LongRAG, and RAPTOR) use the same reranker (\texttt{marco-miniLM}), and LongRAG's reported optimal hyperparameters ($k_1=100$ initial chunks, $k_2=7$ final chunks, unless in ablation). 
We also applied seven different generation methods described in Appendix \ref{app:generation_modes}, and Table \ref{tbl_generation_modes} for detailed comparisons. 

MacRAG-specific parameters are scale-up factor $\alpha \in \{1, 4\}$ and the neighbor hop count $h \in \{0, 1\}$, except during ablation studies. 
The ablation study and performance analysis in Section~\ref{subsec:main_results} demonstrate both robustness and the soundness of the architecture design of MacRAG, enhancing multi-hop reasoning capability.

This controlled setup ensures that performance differences result from MacRAG's architectural innovations, particularly its hierarchical representation and adaptive multi-scale retrieval, rather than from variations in hyperparameters or retrieval components.
By isolating architectural factors, we can clearly assess the impact of multi-scale retrieval on multi-hop reasoning.

\subsection{Main Results}
\label{subsec:main_results}

\begin{table*}[t]
\centering
\scriptsize
\renewcommand{\arraystretch}{1.1}
\setlength{\tabcolsep}{0.2pt}
\centering
\begin{tabular}{l|c c c c c c cc}
\hline
\textbf{($k_2$=7, \textit{marco-miniLM})} & \textbf{R\&B} & \textbf{R\&L} & \textbf{Full\_Ext} & \textbf{Fil} & \textbf{Full\_E\&F} & \textbf{R\&B\_Ext} & \textbf{R\&B\_E\&F} & \textbf{Average} \\
\hline
\rowcolor{middlegrey}
\multicolumn{9}{c}{\textbf{HotpotQA}} \\
LongRAG & 65.46 & 69.40 & 66.60 & 63.32 & 66.20 & 66.08 & 66.09 & 66.16 \\
\textbf{MacRAG}
 & 67.15 (\textbf{+1.69$\uparrow$})
 & 69.55 (\textbf{+0.15$\uparrow$})
 & 69.00 (\textbf{+2.40$\uparrow$})
 & 65.44 (\textbf{+2.12$\uparrow$})
 & 68.52 (\textbf{+2.32$\uparrow$})
 & 66.36 (\textbf{+0.28$\uparrow$})
 & 66.53 (\textbf{+0.44$\uparrow$})
 & \textbf{67.51} (\textbf{+1.35$\uparrow$}, \textbf{+2.04\%$\uparrow$}) \\

\textbf{- Propagation\&Merging}
 & 64.17 (\textbf{-1.29$\downarrow$})
 & 68.91 (\textbf{-0.49$\downarrow$})
 & 67.82 (\textbf{+1.22$\uparrow$})
 & 68.06 (\textbf{+4.74$\uparrow$})
 & 68.44 (\textbf{+2.24$\uparrow$})
 & 63.77 (\textbf{-2.31$\downarrow$})
 & 65.57 (\textbf{-0.52$\downarrow$})
 & \textbf{66.68} (\textbf{+0.52$\uparrow$}, \textbf{+0.79\%$\uparrow$}) \\
 
\textbf{- Scaling up }
 & 64.12 (\textbf{-1.34$\downarrow$})
 & 67.57 (\textbf{-1.83$\downarrow$})
 & 67.33 (\textbf{+0.73$\uparrow$})
 & 64.41 (\textbf{+1.09$\uparrow$})
 & 67.26 (\textbf{+1.06$\uparrow$})
 & 64.22 (\textbf{-1.86$\downarrow$})
 & 63.84 (\textbf{-2.44$\downarrow$})
 & \textbf{65.54} (\textbf{-0.62$\downarrow$}, \textbf{-0.94\%$\downarrow$}) \\
\hline

\rowcolor{middlegrey}
\multicolumn{9}{c}{\textbf{2WikimultihopQA}} \\
LongRAG & 59.97 & 62.37 & 65.35 & 60.68 & 65.89 & 62.08 & 62.47 & 62.69 \\
\textbf{MacRAG}
 & 59.00 (\textbf{-0.97$\downarrow$})
 & 64.87 (\textbf{+2.50$\uparrow$})
 & 68.97 (\textbf{+3.62$\uparrow$})
 & 68.20 (\textbf{+7.52$\uparrow$})
 & 73.19 (\textbf{+7.30$\uparrow$})
 & 66.50 (\textbf{+4.42$\uparrow$})
 & 71.40 (\textbf{+8.93$\uparrow$})
 & \textbf{67.45} (\textbf{+4.76$\uparrow$}, \textbf{+7.59\%$\uparrow$}) \\

\textbf{- Propagation\&Merging} 
 & 59.67 (\textbf{-0.30$\downarrow$})
 & 65.06 (\textbf{+2.69$\uparrow$})
 & 68.48 (\textbf{+3.13$\uparrow$})
 & 67.57 (\textbf{+6.89$\uparrow$})
 & 72.20 (\textbf{+6.31$\uparrow$})
 & 65.13 (\textbf{+3.05$\uparrow$})
 & 72.43 (\textbf{+9.96$\uparrow$})
 & \textbf{67.22} (\textbf{+4.53$\uparrow$}, \textbf{+7.23\%$\uparrow$}) \\

\textbf{- Scaling up}
 & 59.00 (\textbf{-0.97$\downarrow$})
 & 61.89 (\textbf{-0.48$\downarrow$})
 & 67.63 (\textbf{+2.28$\uparrow$})
 & 60.41 (\textbf{-0.27$\downarrow$})
 & 67.03 (\textbf{+1.14$\uparrow$})
 & 66.37 (\textbf{+4.29$\uparrow$})
 & 65.77 (\textbf{+3.30$\uparrow$})
& \textbf{64.01} (\textbf{+1.32$\uparrow$}, \textbf{+2.11\%$\uparrow$}) \\

\hline

\rowcolor{middlegrey}
\multicolumn{9}{c}{\textbf{Musique}} \\
LongRAG & 38.98 & 41.90 & 43.64 & 37.72 & 43.83 & 42.00 & 41.97 & 41.43 \\
\textbf{MacRAG }
 & 44.76 (\textbf{+5.78$\uparrow$})
 & 45.74 (\textbf{+3.84$\uparrow$})
 & 47.42 (\textbf{+3.78$\uparrow$})
 & 47.80 (\textbf{+10.08$\uparrow$})
 & 50.09 (\textbf{+6.26$\uparrow$})
 & 48.57 (\textbf{+6.57$\uparrow$})
 & 51.00 (\textbf{+9.03$\uparrow$})
 & \textbf{47.91} (\textbf{+6.48$\uparrow$}, \textbf{+15.63\%$\uparrow$}) \\

\textbf{- Propagation\&Merging}
 & 41.61 (\textbf{+2.63$\uparrow$})
 & 45.43 (\textbf{+3.53$\uparrow$})
 & 48.00 (\textbf{+4.36$\uparrow$})
 & 47.48 (\textbf{+9.76$\uparrow$})
 & 52.26 (\textbf{+8.43$\uparrow$})
 & 40.09 (\textbf{-1.91$\downarrow$})
 & 45.84 (\textbf{+3.87$\uparrow$})
 & \textbf{45.82} (\textbf{+4.39$\uparrow$}, \textbf{+10.58\%$\uparrow$}) \\

\textbf{- Scaling up }
 & 41.65 (\textbf{+2.67$\uparrow$})
 & 45.87 (\textbf{+3.97$\uparrow$})
 & 46.80 (\textbf{+3.16$\uparrow$})
 & 41.83 (\textbf{+4.11$\uparrow$})
 & 46.59 (\textbf{+2.76$\uparrow$})
 & 39.40 (\textbf{-2.60$\downarrow$})
 & 39.63 (\textbf{-2.34$\downarrow$})
 & \textbf{43.11} (\textbf{+1.68$\uparrow$}, \textbf{+4.04\%$\uparrow$}) \\

\hline
\end{tabular}
\caption{
\label{tbl_experimental_results_ablation_study}
Ablation study of MacRAG with GPT-4o and F1-score, using the same $k_1$, $k_2$, and reranker.
}
\end{table*}

\begin{figure*}[t]
\centering
\includegraphics[width=0.328\textwidth]{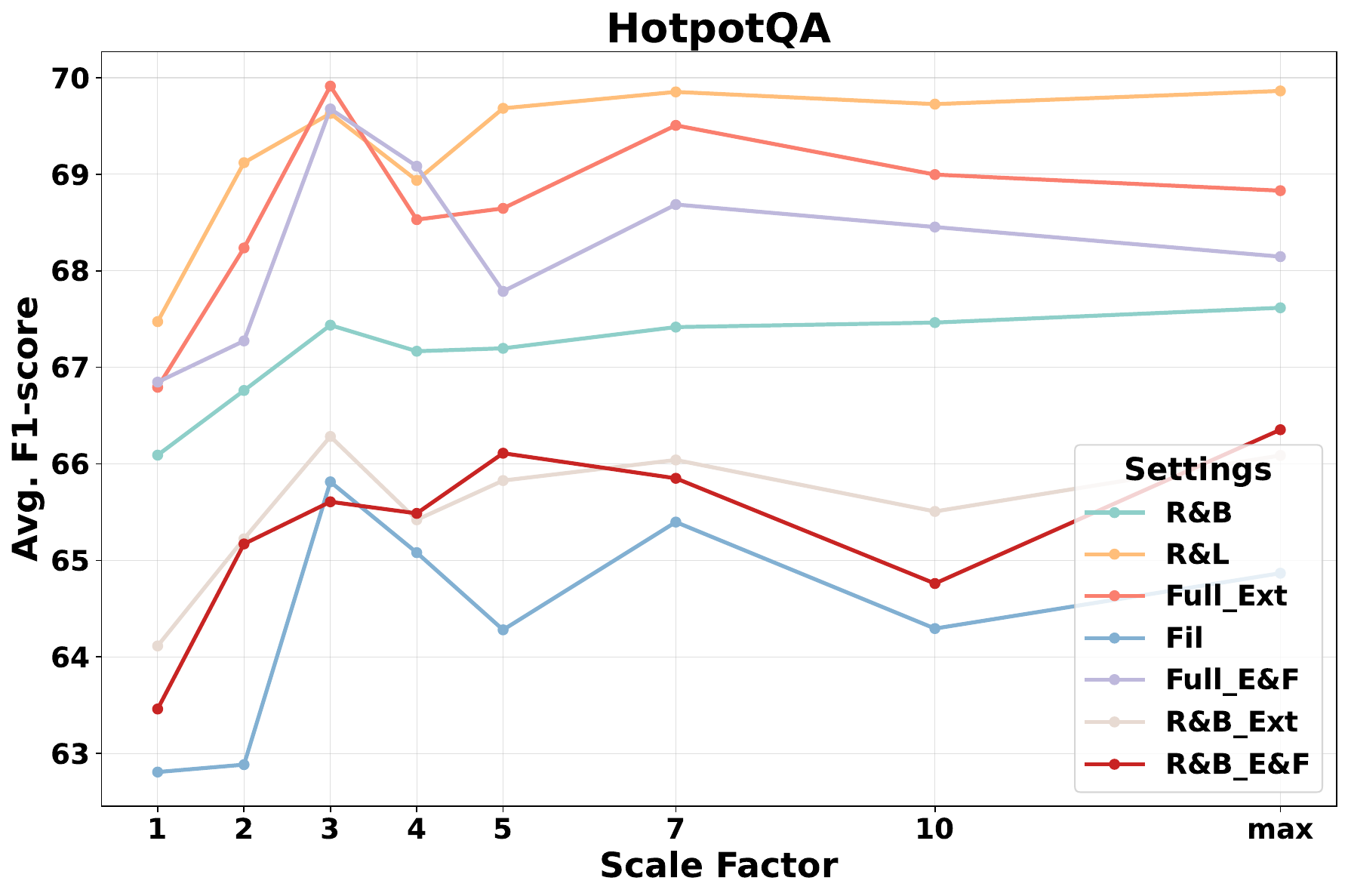}
\includegraphics[width=0.328\textwidth]{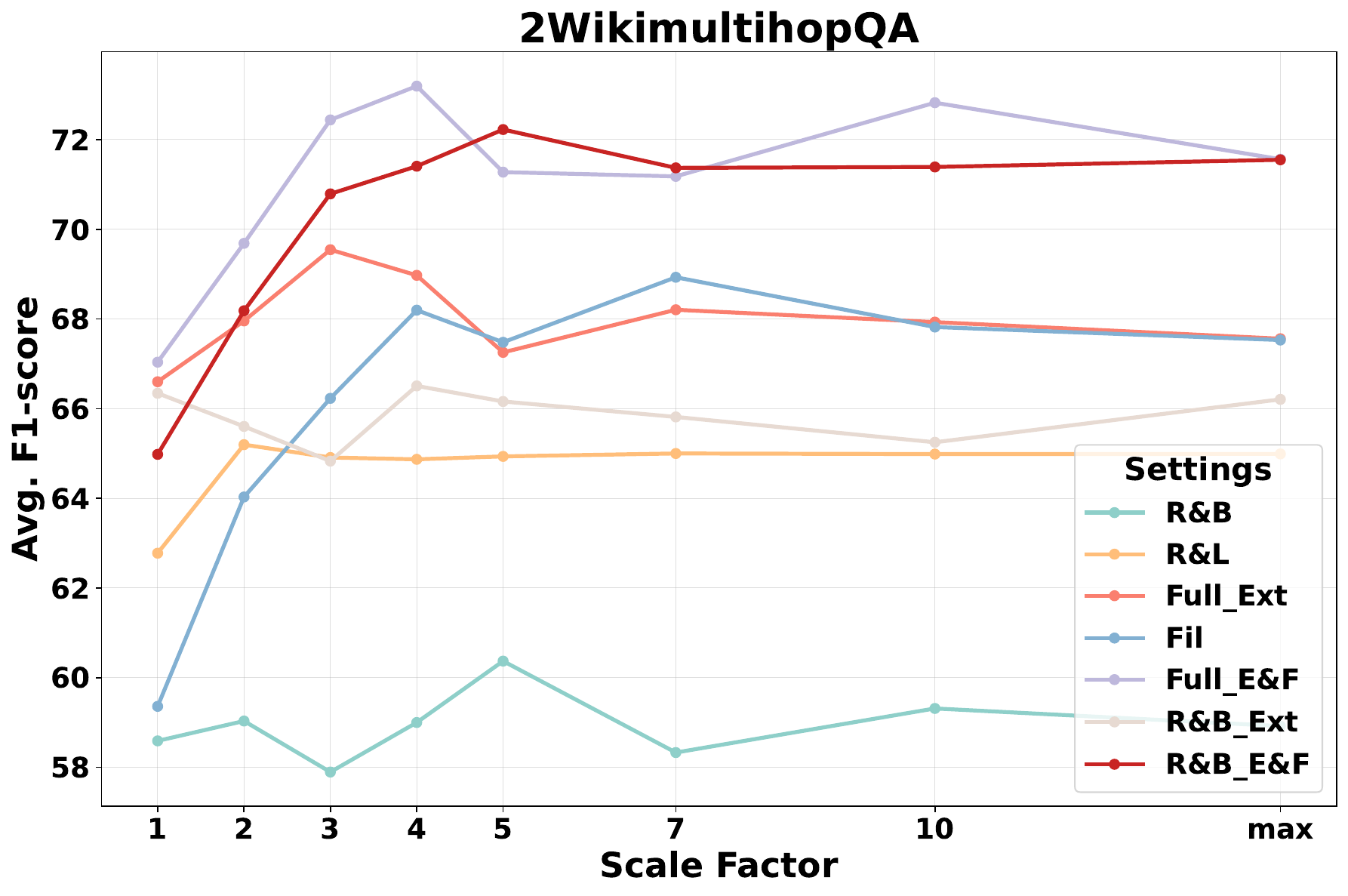}
\includegraphics[width=0.328\textwidth]{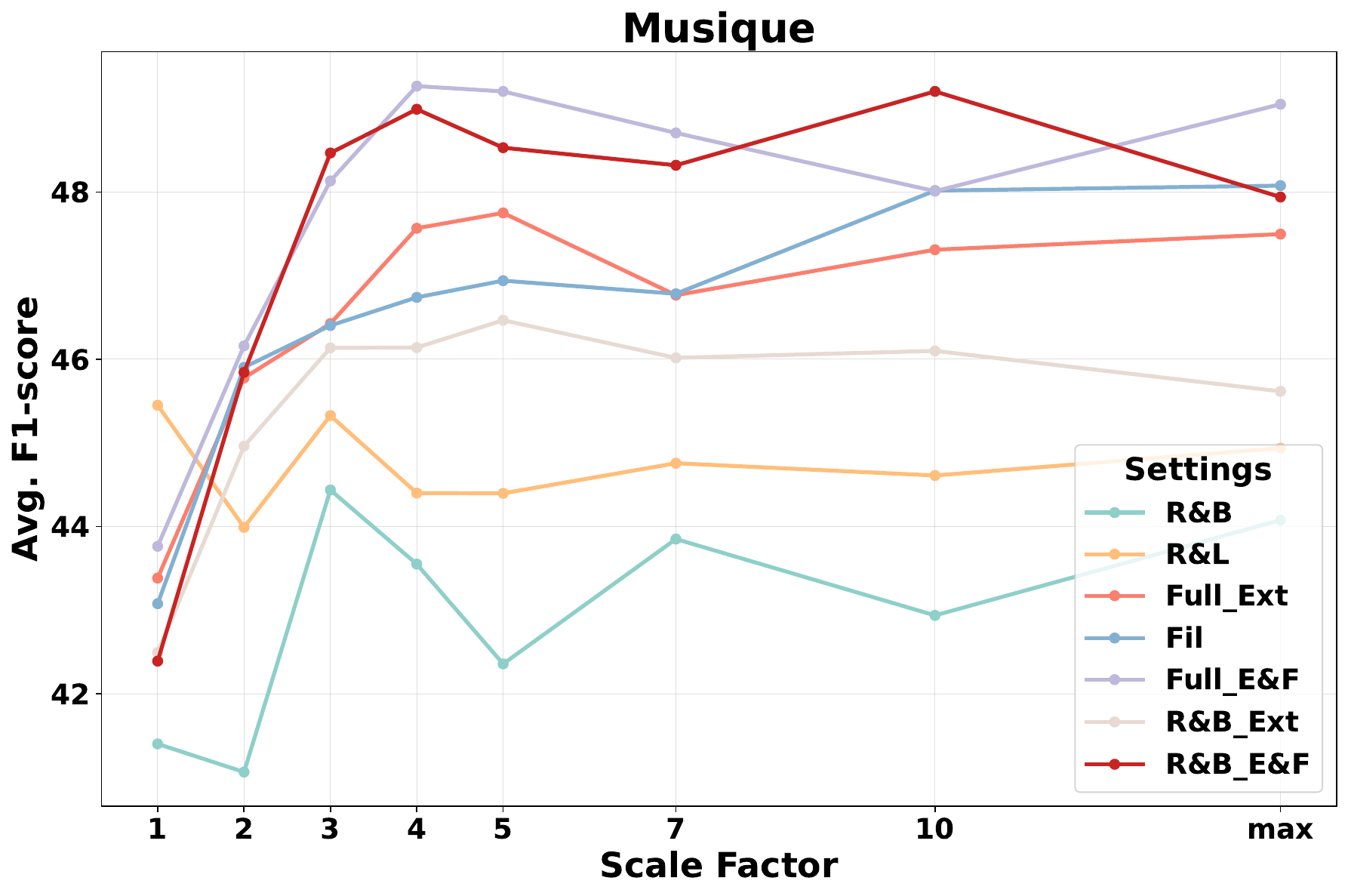}
\caption{Performance trends across datasets for scale factors ($\alpha$).
}
\label{fig_performance_through_scaling_up}
\end{figure*}

\paragraph{Comparison with RAPTOR.}
While RAPTOR \cite{sarthiraptor}, a summarization and hierarchical clustering-based approach, shows gains on standard datasets, it struggles with retrieval quality on LongBench versions \cite{bai2023longbench} as shown in Table \ref{tbl_experimental_results_MacRAG_vs_LongRAG_and_others}. Specifically, RAPTOR with Llama-3.1-8B shows performance decreases in 2WikimultihopQA and Musique, potentially associated with retrieval difficulty on Musique-Normal in \citep{yue2024inference}.
These results highlight the importance of both precise retrieval and effective long-context construction for achieving consistent gains across various multi-hop reasoning and long-context datasets.

\paragraph{Integration with Multi-step Generation Method.}
Integrating MacRAG with the LongRAG (\texttt{E\&F}) generation method across multiple LLMs (Llama-3.1-8B, Gemini-1.5-pro, and GPT-4o) yields substantial performance improvements, as shown in Table \ref{tbl_experimental_results_MacRAG_vs_LongRAG_and_others}. 
This underscores MacRAG’s versatility and robustness as a retrieval approach when paired with the LongRAG generation method.
By constructing an effective long context, MacRAG demonstrates advantages in multi-step generation with LongRAG (\texttt{E\&F}), producing consistent and meaningful gains across three datasets and three LLMs. In contrast, it is not straightforward to extend RAPTOR in combination with LongRAG (\texttt{E\&F}).
As further illustrated in Table \ref{tbl_experimental_results_MacRAG_vs_LongRAG_and_others}, the performance gains for Gemini-1.5-pro and GPT-4o indicate that pairing MacRAG with more powerful LLMs can achieve even greater improvements, despite already high F1-scores obtained by other frameworks under these settings.
We further observe that summarization during indexing improves retrieval quality. Our preliminary experiments show a 7.2\% gain in precision when using summaries instead of raw slices.

\paragraph{Robustness across Generation Methods.}
In addition to MacRAG's robust gains in both vanilla single-step (\texttt{R\&B}) and multi-step generation (\texttt{E\&F}), MacRAG consistently shows strong advantages across the four metrics listed in Table \ref{tbl_average_performances_on_seven_generations}, covering all seven generation modes in Table \ref{tbl_generation_modes}.
These substantial improvements across various datasets, metrics, and LLMs demonstrate the effectiveness of MacRAG’s long-context retrieval in enhancing both precision and recall for RAG. 
Notably, MacRAG achieves significant gains on the Musique datasets, which \cite{yue2024inference} identifies as having challenging retrieval conditions. 
Beyond the average gains in these four metrics, Figure \ref{fig_seven_generation_method_all_datasets_for_Gemini_GPT_f1_score} further confirms MacRAG's robust benefits in all seven single-/multi-step generation schemes in Table \ref{tbl_generation_modes} in Appendix \ref{app:generation_modes}.
Finally, regardless of the choice of $k_2$ or reranker, MacRAG maintains substantial advantages in all test settings in Table \ref{tbl_robust_test_results_MacRAG_vs_LongRAG_gpt4o}.

\paragraph{Enhanced Gains with Stronger Models.}
When applied to the more powerful Gemini-1.5-pro and GPT-4o models, MacRAG showcases its capacity to further improve performance by optimizing the retrieval process and building more relevant, informative contexts. Its use of hierarchical chunking and slicing, coupled with adaptive propagation and merging, maintains high precision while expanding coverage. This leads to more efficient handling of long documents, reduced computational overhead, and higher-quality generated answers overall.

\paragraph{Ablation Study.}
Table \ref{tbl_experimental_results_ablation_study} presents an ablation study demonstrating the effectiveness of MacRAG's key components: Propagation and Merging, Scaling-up, and Hierarchical Slicing Retrieval. 
Each component contributes to cumulative performance gains.
The combination of propagation and merging with the scaling-up mechanism increases the coverage of relevant contexts for multi-hop reasoning tasks. Figure \ref{fig_performance_through_scaling_up} further illustrates the benefits of scaling up promising candidates with $\alpha= [2, 3, 4]$, showing how this approach effectively balances context lengths while incorporating additional relevant candidates.
We observe that performance gains remain consistent across different $k_2$ values and rerankers, indicating robustness.
Although the parameters $(k_1, k_2, \alpha, h)$ are fixed, MacRAG adaptively merges nearby slices depending on partial relevance signals. Removing this adaptivity leads to a 5\% F1 drop, as shown in our ablations.

\subsection{Generation Schemes and Input Lengths}
\label{subsec:generation_input_lengths}
\begin{figure}[t]
\centering
\includegraphics[width=\linewidth]{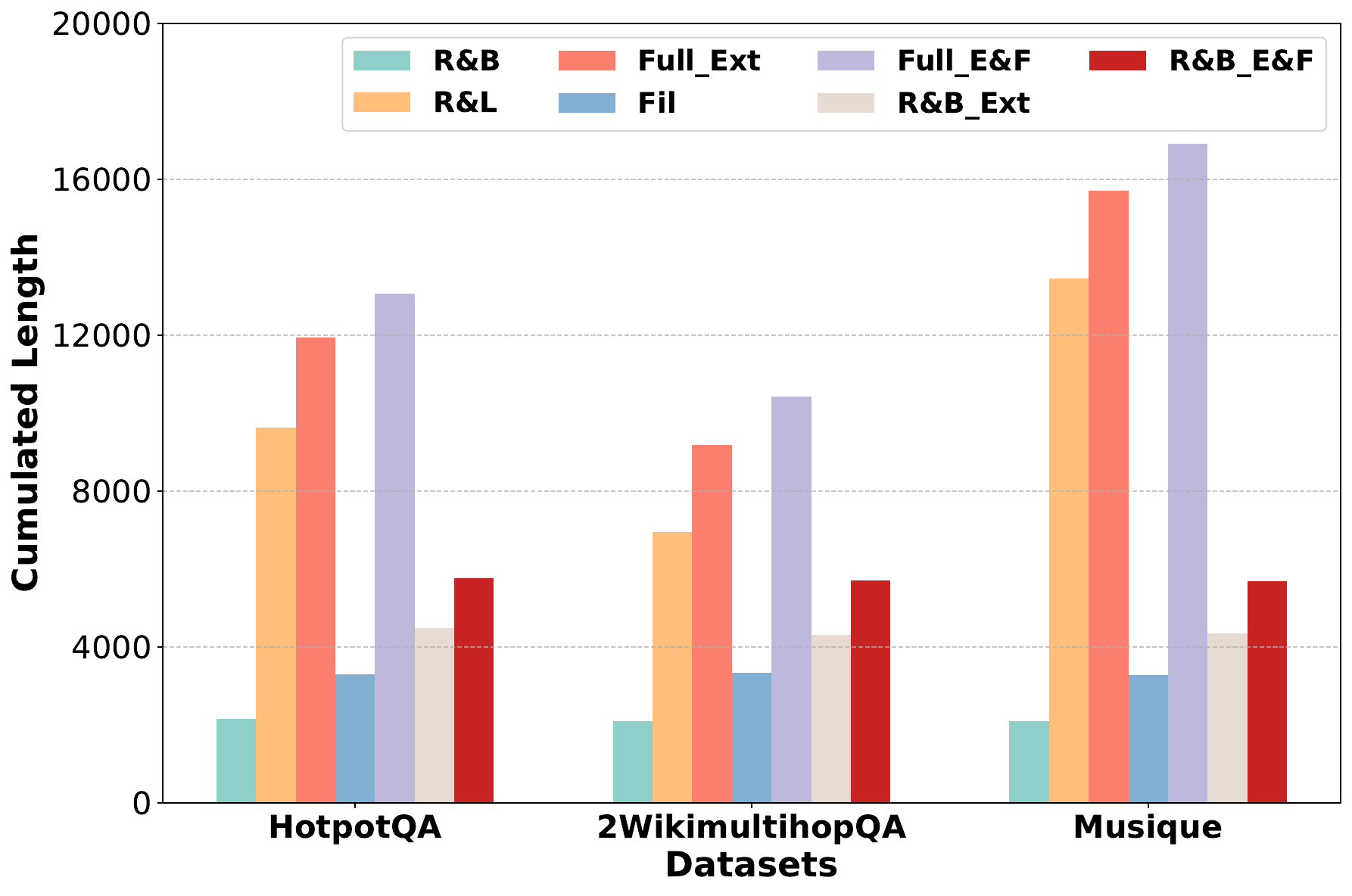}    
\caption{
Cumulative LLM's input context lengths for single/multi-step generation methods in Table \ref{tbl_generation_modes}.}
\label{fig_Cumulative_Input_Context_Length}
\end{figure}
While multi-step generation methods (e.g., LongRAG's Full\_E\&F) improve answer quality, they often increase cumulative input length. Figure~\ref{fig_Cumulative_Input_Context_Length} compares cumulative input lengths for single and multi-step generation methods in Table~\ref{tbl_generation_modes}, given single retrieval with specific settings, highlighting that \texttt{R\&L}, \texttt{Full\_Ext}, and \texttt{Full\_E\&F} accumulate the most overhead by relying on entire documents. 
In contrast, MacRAG’s final $k_2$ chunks, though potentially larger than basic R\&B chunks, preserve bounded context well below full-document with long lengths.

As Figure~\ref{fig_seven_generation_method_all_datasets_for_Gemini_GPT_f1_score} and Table~\ref{tbl_experimental_results_ablation_study} illustrate, MacRAG consistently outperforms baselines under both \texttt{R\&B\_Ext} and \texttt{R\&B\_E\&F} settings. 
Due to MacRAG's advantage in effective and stable long context construction, even our intermediate \texttt{R\&B\_E\&F} variant demonstrates significant improvements as shown in Table \ref{tbl_experimental_results_ablation_study}. 
For example, on 2WikiMultihopQA (GPT-4o), MacRAG (\texttt{R\&B\_E\&F}) achieves an F1 of 72.43 versus LongRAG (\texttt{Full\_E\&F}) at 65.89 (+9.93\% relative gain), while using 8.19\% less context. 
On Musique, the intermediate variant of MacRAG (\texttt{R\&B\_E\&F}) reaches 45.84 compared to LongRAG (\texttt{Full\_E\&F})'s 43.83 (+4.58\% relative) with a 45.87\% reduction in context length. 
This highlights MacRAG's efficiency in evidence gathering and context construction.

\subsection{Efficiency and Latency Analysis}
\label{subsec:efficiency}

MacRAG maintains sub-second efficiency despite constructing richer multi-scale contexts.  
On average, retrieval and reranking take 0.23 seconds per query, which is 38\% faster than RAPTOR’s 0.37 seconds.  
With Llama 3.1 8B, total inference latency is 0.99 seconds (0.23s retrieval and 0.76s generation), compared to 0.80 seconds for RAPTOR (0.37s and 0.43s, respectively).  
This slight increase in generation time is offset by improved retrieval precision and higher answer quality on R\&B queries.

This efficiency is enabled by MacRAG’s lightweight, index-based context construction and shared reranking modules, which preserve the same level of computational complexity as baseline retrieval pipelines.  
MacRAG introduces a one-time offline summarization step during indexing, similar to RAPTOR and SIRERAG.  
This step runs in parallel with standard RAG indexing, adding no runtime latency and supporting efficient index maintenance.  
As shown in Table~\ref{tbl_experimental_results_ablation_study}, Figure~\ref{fig_Cumulative_Input_Context_Length}, and Subsection~\ref{subsec:generation_input_lengths},  
MacRAG achieves a better trade-off than LongRAG between input context length, answer quality, and overall computational cost.

\section{Conclusion}
In this work, we introduced Multi-scale Adaptive Context RAG (MacRAG), a novel framework designed to address key trade-offs in RAG involving precision, coverage, and computational efficiency for long-context question answering and multi-hop reasoning.  
By dynamically constructing query-adaptive long contexts through real-time, multi-scale retrieval, MacRAG demonstrably improves both precision and recall while maintaining operational efficiency.
Our extensive experiments validated MacRAG's advantages across diverse datasets and both single- and multi-step generation paradigms, showcasing consistent and notable performance gains.
These findings establish MacRAG as a potent and efficient core module for advancing next-generation iterative and agentic RAG systems, with significant potential for robust enterprise-level applications.
\section*{Limitations}
While MacRAG demonstrates strong performance with its modular and efficient design, several avenues offer potential for further extension. The current offline summarization and indexing strategy, though effective and its costs amortized, could benefit from adaptive or online variants to better handle rapidly evolving corpora. Furthermore, while the use of fixed hyperparameters such as chunk size and expansion scale has shown robustness, dynamically adjusting these parameters based on query complexity presents a promising research direction. Finally, tighter integration of MacRAG with sophisticated agentic or multi-round retrieval workflows could further enhance its applicability in highly interactive and real-time RAG scenarios.


\bibliography{main}

\clearpage
\appendix
\section{Appendix}
\label{appendix}

\subsection{Generation Modes}
\label{app:generation_modes}
Table~\ref{tbl_generation_modes} summarizes different generation modes. 
In addition to MacRAG's robust gains in both vanilla single-step (\texttt{R\&B}) and multi-step generation (\texttt{E\&F}), MacRAG consistently shows strong advantages across the four metrics listed in Table \ref{tbl_average_performances_on_seven_generations}, covering all seven generation modes in Table \ref{tbl_generation_modes}.
These substantial improvements across various datasets, metrics, and LLMs demonstrate the effectiveness of MacRAG’s long-context retrieval in enhancing both precision and recall for RAG. 
Notably, MacRAG achieves significant gains on the Musique datasets, which \cite{yue2024inference} identifies as having challenging retrieval conditions. 
Beyond the average gains in these four metrics, Figure \ref{fig_seven_generation_method_all_datasets_for_Gemini_GPT_f1_score} further confirms MacRAG's robust benefits in all seven single-/multi-step generation schemes in Table \ref{tbl_generation_modes} in Appendix \ref{app:generation_modes}.
Finally, regardless of the choice of $k_2$ or reranker, MacRAG maintains substantial advantages in all test settings in Table \ref{tbl_robust_test_results_MacRAG_vs_LongRAG_gpt4o}.

\begin{table*}[ht]
\centering
\footnotesize
\begin{tabular}{@{}p{1cm}p{14.5cm}@{}}
\toprule
\textbf{Mode} & \textbf{Description} \\
\midrule
\textbf{\texttt{R\&B}} & Single-step generation using retrieved top-$k_2$ chunks, which is the vanilla basic version.\\
\textbf{\texttt{R\&L}} & Single-step generation using complete documents from top-$k_2$ chunks. Prone to the ``Lost in the Middle'' effect and inefficient for long documents. \\
\textbf{\texttt{Full\_Ext}} & Multi-step generation: (1) LLM extracts query-relevant content from full documents of top-$k_2$ chunks, (2) generates answer using both chunks and extracted content.  \\
\textbf{\texttt{Fil}} & Filtering re-ranked top-$k_2$ chunks, \textit{e.g.}, via LLM. Improves precision but may lose borderline info. \\
\textbf{\texttt{Full\_E\&F}} & Multi-step generation combining \texttt{Full\_Ext}'s extracted content and \texttt{Fil}'s filtered chunks. Balances coverage and precision, but is expensive for long documents. \\
\textbf{\texttt{R\&B\_Ext}} & Multi-step generation with content extraction limited to top-$k_2$ chunks, reducing computational overhead. \\
\textbf{\texttt{R\&B\_E\&F}} & Multi-step generation combining \texttt{R\&B\_Ext}'s extracted content and \texttt{Fil}'s filtered chunks. Optimizes precision-recall trade-off w/o processing the full document. \\
\bottomrule
\end{tabular}
\caption{
\label{tbl_generation_modes}
Seven different generation methods with single-step and multi-step generation schemes using retrieved top-$k_2$ chunks.
}
\end{table*}

\subsection{Benchmark Datasets and Challenges}
\label{sec:benchmark_datasets}
The LongBench variants of {HotpotQA}, {2WikimultihopQA}, and {Musique} datasets have become the standard evaluation suite for hierarchical retrieval research, including recent systems such as SIRERAG \cite{zhang2025sirerag}, RAPTOR \cite{sarthiraptor}, and LongRAG \cite{zhao2024longrag}.
They specifically intensify the ``Lost in the Middle" phenomenon by introducing near-duplicate and tangential passages, creating challenging retrieval environments where critical connecting information becomes obscured by surrounding context.
Each dataset introduces distinct multi-hop reasoning challenges that systematically test different aspects of retrieval capability:
HotpotQA features bridging and comparison queries requiring evidence synthesis across multiple paragraphs.
2WikiMultihopQA expands partial excerpts to complete articles, significantly increasing the risk that essential connecting information becomes buried within expansive context.
Musique embeds nested sub-questions within tangential sections, challenging systems to identify and connect scattered bridging facts across document boundaries.
These characteristics directly test a retrieval system's ability to handle fragmented evidence, reconcile overlapping information, and identify critical bridging facts despite contextual noise \citep{yue2024inference,leng2024long}, precisely the challenges that MacRAG's multi-scale architecture addresses through hierarchical indexing and adaptive retrieval.

\subsection{Target Baseline Methods}
\label{sec:baseline_methods}
Based on SIRERAG's comparative analysis (Table 4 in \citet{zhang2025sirerag}), we focus our evaluation on RAPTOR \cite{sarthiraptor} and LongRAG \cite{zhao2024longrag} as representative state-of-the-art hierarchical RAG systems. 
SIRERAG's analysis establishes that GraphRAG underperforms on these specific multi-hop tasks, while RAPTOR, HippoRAG, and SIRERAG achieve comparable performance in terms of F1-score on multiple multi-hop question answering datasets \cite{zhang2025sirerag}.
Among these top-performing systems, we selected RAPTOR and LongRAG because they balance performance and computational efficiency better, offering lower computational costs and easier maintenance than SIRERAG's higher overhead.

\subsection{Adaptive Retrieval Mechanism Analysis}
A key strength of MacRAG lies in its adaptive retrieval mechanism, which dynamically constructs query-specific contexts despite using fixed common hyperparameters ($k_1$, $k_2$) and MacRAG's $\alpha$, and $h$. 
This adaptivity operates through several complementary mechanisms:



\paragraph{Query-dependent Adaptive Context Expansion:} For complex multi-hop queries, MacRAG naturally extends its context boundaries to encompass connecting information. 
For simpler queries, expansion remains more focused, selectively around the most promising content regions, but avoiding unnecessary content.
Our ablation studies quantify the impact of this adaptivity. 
When the adaptivity of query-adaptive expansion of neighborhoods is removed by eliminating the propagation and merging steps (Table \ref{tbl_experimental_results_ablation_study}), performance drops by up to around 5\% F1-score on datasets. 
This degradation is most pronounced on Musique, where connecting multiple scattered pieces of evidence is crucial.
Table \ref{tbl_experimental_results_ablation_study} additionally reveals another important aspect of query-adaptive expansion of borderline candidates "Scaling-up", and if we eliminate its corresponding "Scaling-up" component, then performance drops up to around 6\% F1-score on datasets.  
This aligns with our expectation that adaptive expansion becomes increasingly valuable as query complexity grows.

\subsection{Preprocessing Efficiency of MacRAG}
\label{sec:appendix_efficiency_MacRAG}

\paragraph{Chunk Size and Token Budget:}
MacRAG uses original chunk sizes of approximately 400 tokens (around 1500 characters), which fall within the 200--500 token range used in prior work such as LongRAG even for re-ranking on retrieved chunks.
The average sizes of summarized chunks are around 800-1000 characters.
We applied slices into summarized chunks for initial retrieval to improve precision further with two slice sizes around 450 characters with 300 character overlaps or 600 characters with 450 character overlaps.
These sizes allow MacRAG to maintain sub-second retrieval and reranking times while keeping the context structure manageable for downstream language models.

\paragraph{Offline Summarization Cost:}
MacRAG includes a one-time summarization step during indexing, which does not affect query-time latency. This step is comparable in cost to offline embedding computations in standard RAG pipelines. All runtime efficiency measurements reported here exclude this indexing cost, as it is amortized over the use of the document collection.

\subsection{Latency Comparison between MacRAG and RAPTOR}
\label{sec:appendix_efficiency_comparison_to_RAPTOR}

\paragraph{Retrieval and Reranking Latency:}
MacRAG maintains sub-second latency for retrieval and reranking across all evaluated datasets. Using dense retrieval with $k_1 = 100$ followed by cross-encoder reranking with $k_2 = 7$, the measured latency is 0.23 seconds on HotpotQA, 0.22 seconds on 2WikimultihopQA, and 0.24 seconds on Musique. 
In contrast, RAPTOR reports 0.43s, 0.24s, and 0.44s on the same datasets, averaging 0.37s. This represents a 38\% speedup on average, even though both methods operate on the same top-100 retrieved chunks and use the same reranker (marco-miniLM). 
MacRAG achieves this by leveraging index-based merging after reranking, which reuses precomputed relevance scores and adds only a few milliseconds per query.

\paragraph{Generation Time and Trade-offs:}
While retrieval and reranking are efficient, generation time varies depending on the context length produced by each method. MacRAG’s adaptive multi-scale expansion typically results in longer but more relevant contexts. With Llama 3.1 8B, the average generation time with \texttt{R\&B} is 0.76 seconds, compared to 0.43 seconds for RAPTOR with \texttt{R\&B}. This increase reflects MacRAG's strategy of incorporating broader evidence to improve answer accuracy. Users seeking faster inference can reduce the expansion parameter $\alpha$ to trade off coverage for speed, though at the cost of some performance.

\paragraph{Total Latency Comparison:}
Combining retrieval, reranking, and generation with \texttt{R\&B}, \textbf{the total end-to-end latency of MacRAG is 0.99 seconds (0.23s + 0.76s), compared to RAPTOR’s 0.80 seconds (0.37s + 0.43s)}. 
While MacRAG incurs a small additional cost, it consistently yields higher accuracy across multiple datasets and models, as shown in Table~\ref{tbl_experimental_results_MacRAG_vs_LongRAG_and_others}.

\subsection{Efficiency Comparison between MacRAG and LongRAG}
\label{sec:appendix_efficiency_comparison_to_LongRAG}
MacRAG and LongRAG both rely on index-based hierarchical retrieval and reranking over the same number of $k_1$ chunks. 
As a result, their retrieval and reranking steps show similar efficiency and latency.  
However, for the generation step, according to Table~\ref{tbl_experimental_results_ablation_study}, Figure~\ref{fig_Cumulative_Input_Context_Length}, and Subsection~\ref{subsec:generation_input_lengths}, MacRAG achieves improved efficiency compared to LongRAG in terms of the trade-off between final answer quality, input context length, and overall computational cost.

\subsection{Discussion on Effectiveness of MacRAG's Retrieval and Constructed Long Context}
\label{subsec:discussion}

\textbf{Enhanced Precision and Coverage:~}
MacRAG employs a hierarchical multi-scale strategy that begins with fine-grained slice retrieval and chunk-level re-ranking for precise identification of relevant content, then systematically incorporates broader document-level context through $h$-hop neighbor expansions. 
To preserve coherence without overwhelming the LLM, it strategically up-scales via the $(\alpha \times k_2)$ factor, capturing borderline yet crucial segments. Rather than relying on compressed text or entire documents, MacRAG focuses on real-time construction of effective long contexts from promising original chunks while enforcing an upper bound on context length. 
By removing irrelevant portions from documents and maintaining a continuous, coherent subset of text, MacRAG avoids excessive token consumption, mitigates the ``lost in the middle" phenomenon, and minimizes hallucinations, ultimately ensuring high recall for complex multi-hop queries.

\subsection{Discussion on Single, Multi-Step, Iterative, and Agentic Generation with MacRAG}
\label{subsec:gen8}
MacRAG's modular design seamlessly supports single and multi-step generation, iterative retrieval, and agentic pipelines. 
By combining its core multi-scale retrieval with standard generation methods, MacRAG dynamically adapts context formation and refinement across multiple rounds without overwhelming the LLM. 
In single and multi-step modes (\textit{e.g.}, LongRAG~\cite{zhao2024longrag} and standard multi-step QA in Section~\ref{subsec_generation_modes}), MacRAG selects and assembles relevant text from large corpora while minimizing noise and retaining crucial connections, thereby improving both recall and precision for multi-hop reasoning. 
For iterative scenarios, such as IterDRAG~\cite{yue2024inference} or chain-of-RAG~\cite{wang2025chainofretrievalaugmentedgeneration}, MacRAG can support efficient updates on the evidence set via
$ \mathcal{C}_{t} \leftarrow \mathrm{Merge}\Bigl(\mathcal{C}_{t-1},\, \mathrm{Retrieval}(q_{t})\Bigr) $
to unify old and newly retrieved content, promoting consistent coverage of bridging facts while discarding irrelevant material. 
In agentic RAG settings~\cite{asai2024selfrag,jeong2024adaptive, chen2025improvingretrievalaugmentedgenerationmultiagent}, MacRAG likewise prevents context explosion by focusing specifically on segments required for each action, thereby improving precision and recall over multiple steps. 
Our experiments (Section~\ref{experiments}) confirm that integrating MacRAG with single and multi-step generation methods, potential core modules for multi-round advanced systems, consistently enhances complex multi-hop reasoning tasks and reduces error rates in long-context RAG.

\subsection{Discussion on Graph-Enhanced MacRAG}
A promising direction is extending MacRAG with a two-stage reranking strategy incorporating graph-based knowledge structures. After initial chunk-level reranking, MacRAG could perform a second reranking phase by expanding top candidates through both local index-based extensions and pre-constructed graph neighbors, enabling efficient coverage of both local and global relationships. This approach would leverage MacRAG's bounded context guarantees at each phase while allowing controlled exploration of knowledge-guided connections, managing computational efficiency through bounded candidate sets during reranking. By applying MacRAG's document-oriented indexing or relationship-based thresholding to merge expanded candidates, this strategy would enhance retrieval quality for complex queries requiring both precise local context and broad knowledge integration.

\begin{table*}[t]
\centering
\scriptsize
\renewcommand{\arraystretch}{1.1}
\setlength{\tabcolsep}{0.2pt}
\begin{tabular}{l c c c c c c c c c c}
\hline
\textbf{Model} & $\mathbf{k_2}$ & \textbf{Reranker} &
\textbf{R\&B} & \textbf{R\&L} & \textbf{Full\_Ext} & \textbf{Fil} &
\textbf{Full\_E\&F} & \textbf{R\&B\_Ext} & \textbf{R\&B\_E\&F} & \textbf{Average} \\
\hline
\rowcolor{middlegrey}
\multicolumn{11}{c}{\textbf{HotpotQA}} \\
LongRAG & 7  & bge-m3
 & 67.67 & 67.99 & 68.96 & 64.30 & 68.49 & 66.73 & 66.10 & 67.18 \\
\textbf{MacRAG} & 7 & bge-m3
 & 67.59 \,(\textbf{-0.08$\downarrow$})
 & 68.63 \,(\textbf{+0.64$\uparrow$})
 & 70.53 \,(\textbf{+1.57$\uparrow$})
 & 65.56 \,(\textbf{+1.26$\uparrow$})
 & 70.29 \,(\textbf{+1.80$\uparrow$})
 & 66.59 \,(\textbf{-0.14$\downarrow$})
 & 67.32 \,(\textbf{+1.22$\uparrow$})
 & \textbf{68.07} \,(\textbf{+0.89$\uparrow$, +1.32\%$\uparrow$}) \\
LongRAG & 12 & bge-m3
 & 68.57 & 67.65 & 70.31 & 64.74 & 70.14 & 67.63 & 67.49 & 68.08 \\
\textbf{MacRAG} & 12 & bge-m3
 & 67.88 \,(\textbf{-0.69$\downarrow$})
 & 69.87 \,(\textbf{+2.22$\uparrow$})
 & 70.72 \,(\textbf{+0.41$\uparrow$})
 & 65.98 \,(\textbf{+1.24$\uparrow$})
 & 70.66 \,(\textbf{+0.52$\uparrow$})
 & 69.05 \,(\textbf{+1.42$\uparrow$})
 & 67.84 \,(\textbf{+0.35$\uparrow$})
 & \textbf{68.86} \,(\textbf{+0.78$\uparrow$, +1.15\%$\uparrow$}) \\
\hline
\rowcolor{middlegrey}
\multicolumn{11}{c}{\textbf{2WikimultihopQA}} \\
LongRAG & 7  & bge-m3
 & 59.36 & 65.56 & 68.27 & 55.31 & 67.36 & 64.42 & 63.88 & 64.45 \\
\textbf{MacRAG} & 7 & bge-m3
 & 62.32 \,(\textbf{+2.96$\uparrow$})
 & 66.34 \,(\textbf{+0.78$\uparrow$})
 & 71.63 \,(\textbf{+3.36$\uparrow$})
 & 69.72 \,(\textbf{+14.41$\uparrow$})
 & 73.98 \,(\textbf{+6.62$\uparrow$})
 & 66.61 \,(\textbf{+2.19$\uparrow$})
 & 72.63 \,(\textbf{+8.75$\uparrow$})
 & \textbf{69.03} \,(\textbf{+4.58$\uparrow$, +7.11\%$\uparrow$}) \\
LongRAG & 12 & bge-m3
 & 60.08 & 66.77 & 69.28 & 58.50 & 69.39 & 64.90 & 65.28 & 64.89 \\
\textbf{MacRAG} & 12 & bge-m3
 & 64.29 \,(\textbf{+4.21$\uparrow$})
 & 67.20 \,(\textbf{+0.43$\uparrow$})
 & 70.95 \,(\textbf{+1.67$\uparrow$})
 & 69.46 \,(\textbf{+10.96$\uparrow$})
 & 73.90 \,(\textbf{+4.51$\uparrow$})
 & 67.49 \,(\textbf{+2.59$\uparrow$})
 & 71.80 \,(\textbf{+6.52$\uparrow$})
 & \textbf{69.30} \,(\textbf{+4.41$\uparrow$, +6.80\%$\uparrow$}) \\
\hline
\rowcolor{middlegrey}
\multicolumn{11}{c}{\textbf{Musique}} \\
LongRAG & 7  & bge-m3
 & 42.34 & 48.08 & 47.88 & 42.17 & 47.92 & 43.70 & 44.06 & 45.16 \\
\textbf{MacRAG} & 7 & bge-m3
 & 45.54 \,(\textbf{+3.20$\uparrow$})
 & 46.68 \,(\textbf{-1.40$\downarrow$})
 & 49.58 \,(\textbf{+1.70$\uparrow$})
 & 46.76 \,(\textbf{+4.59$\uparrow$})
 & 49.53 \,(\textbf{+1.61$\uparrow$})
 & 47.94 \,(\textbf{+4.24$\uparrow$})
 & 49.14 \,(\textbf{+5.08$\uparrow$})
 & \textbf{47.88} \,(\textbf{+2.72$\uparrow$, +6.02\%$\uparrow$}) \\
LongRAG & 12 & bge-m3
 & 41.53 & 46.96 & 49.58 & 41.60 & 49.57 & 46.76 & 46.33 & 46.05 \\
\textbf{MacRAG} & 12 & bge-m3
 & 46.44 \,(\textbf{+4.91$\uparrow$})
 & 45.81 \,(\textbf{-1.15$\downarrow$})
 & 51.02 \,(\textbf{+1.44$\uparrow$})
 & 46.70 \,(\textbf{+5.10$\uparrow$})
 & 51.54 \,(\textbf{+1.97$\uparrow$})
 & 50.25 \,(\textbf{+3.49$\uparrow$})
 & 49.50 \,(\textbf{+3.17$\uparrow$})
 & \textbf{48.75} \,(\textbf{+2.70$\uparrow$, +5.86\%$\uparrow$}) \\
\hline
\end{tabular}
\vspace{-2mm}
\caption{
\label{tbl_robust_test_results_MacRAG_vs_LongRAG_gpt4o}
Robust test of MacRAG with various $k_2$ an alternative re-ranker on three multi-hop QA datasets, using GPT-4o and F1-score. The experiments were conducted with the same hyperparameters as reported in LongRAG~\cite{zhao2024longrag}.}
\end{table*}

\begin{figure*}[t!]
\centering
\includegraphics[width=0.24\textwidth]{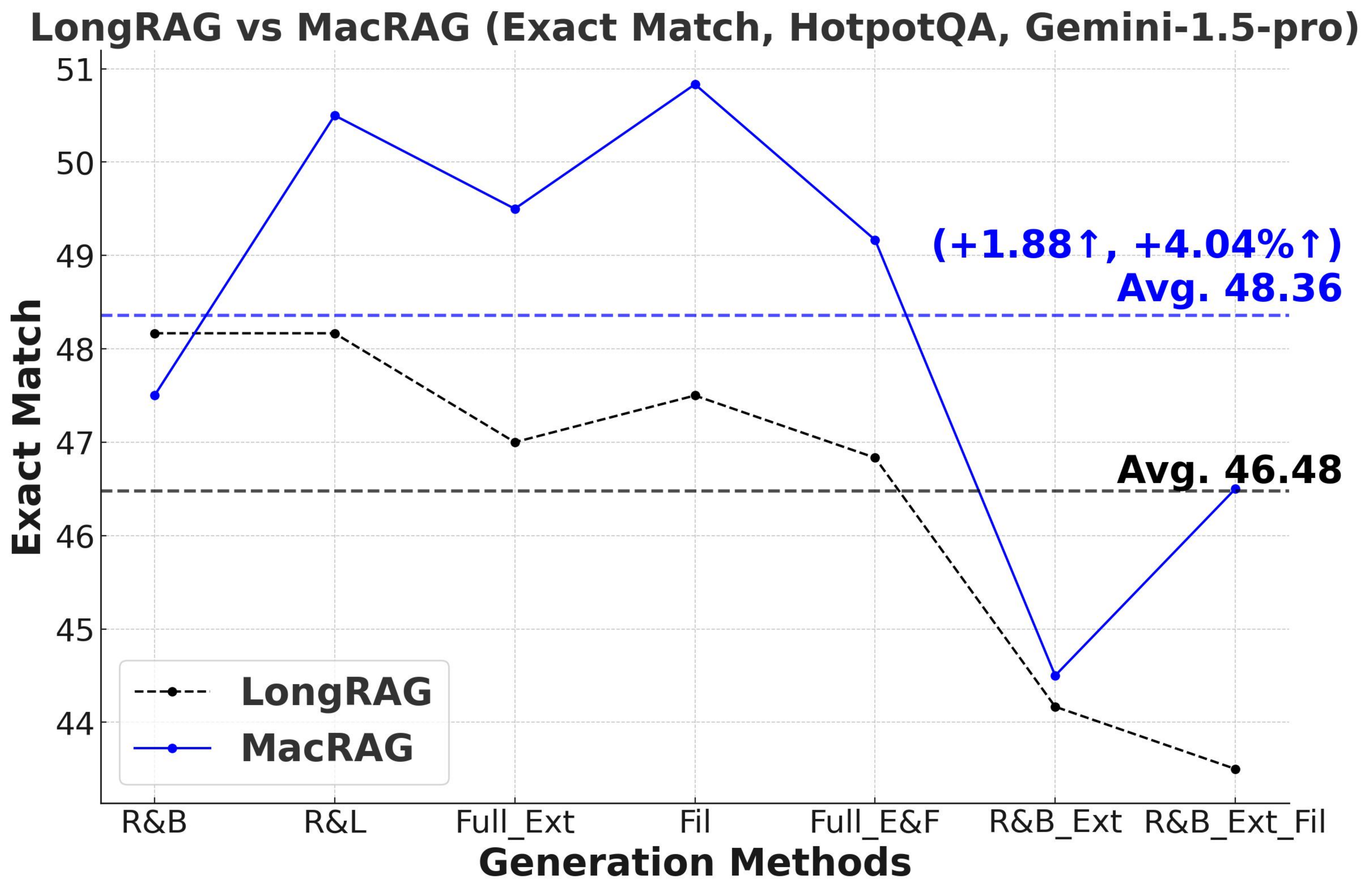}
\includegraphics[width=0.24\textwidth]{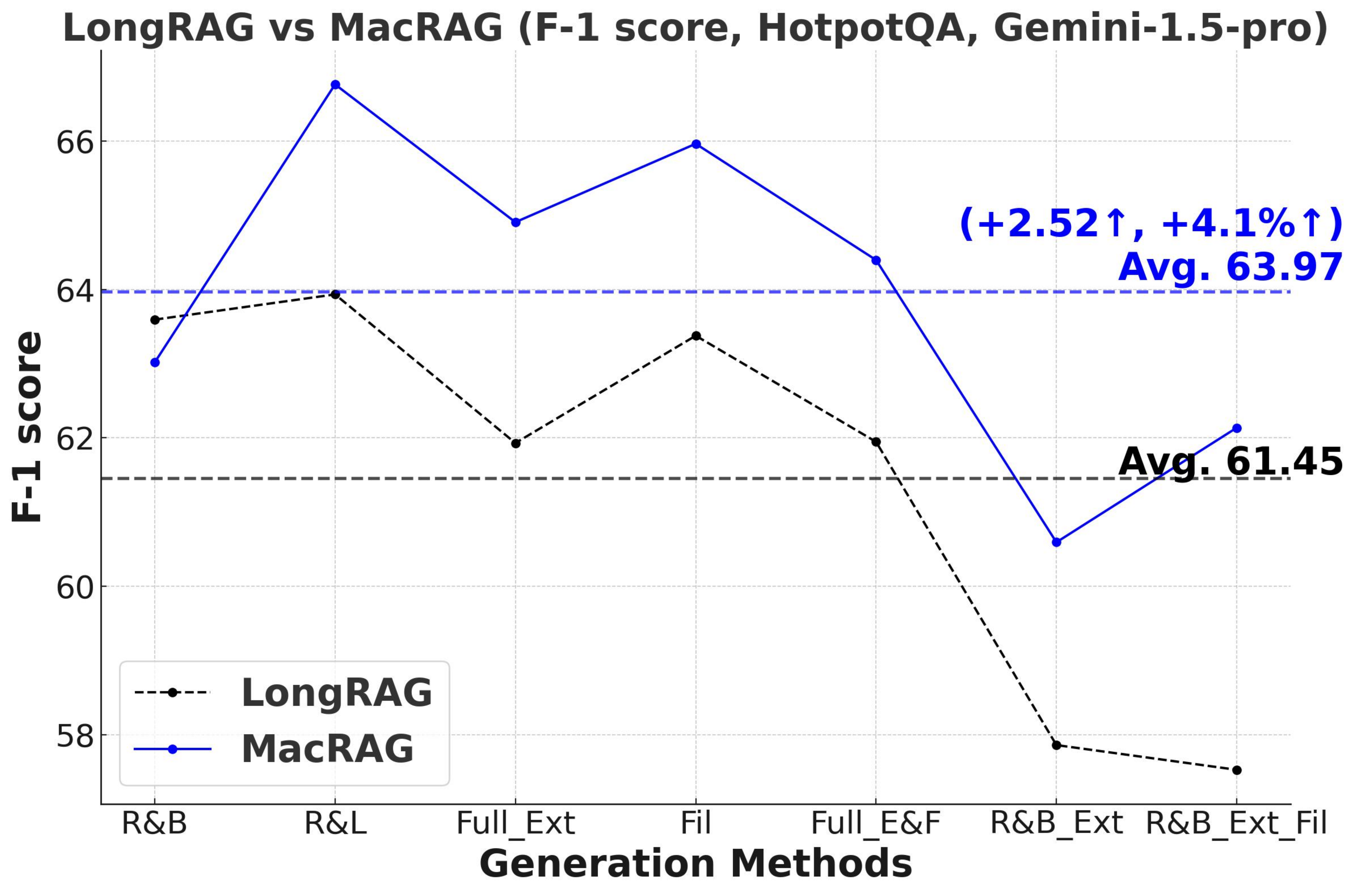}
\includegraphics[width=0.24\textwidth]{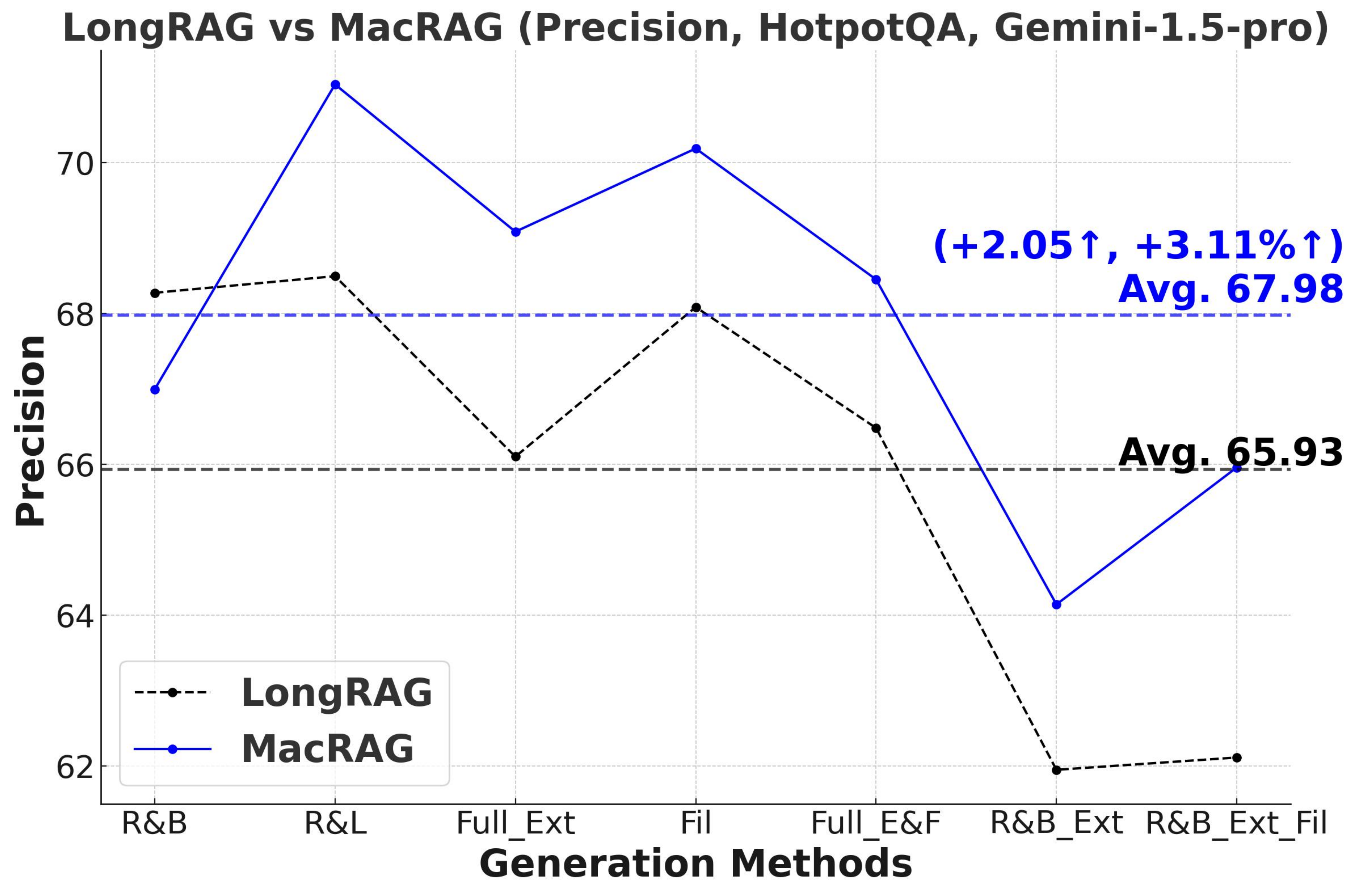}
\includegraphics[width=0.24\textwidth]{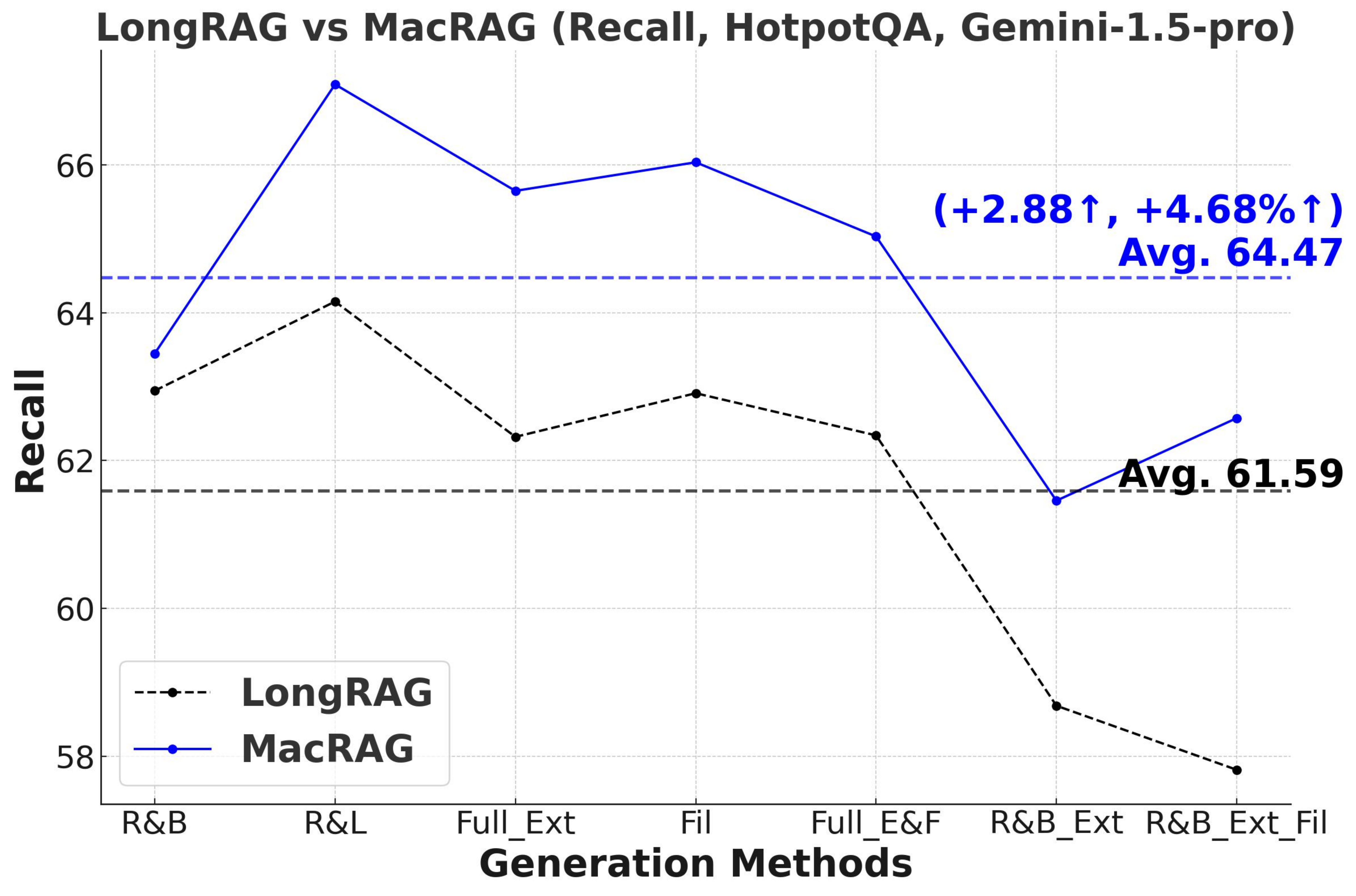}

\includegraphics[width=0.24\textwidth]{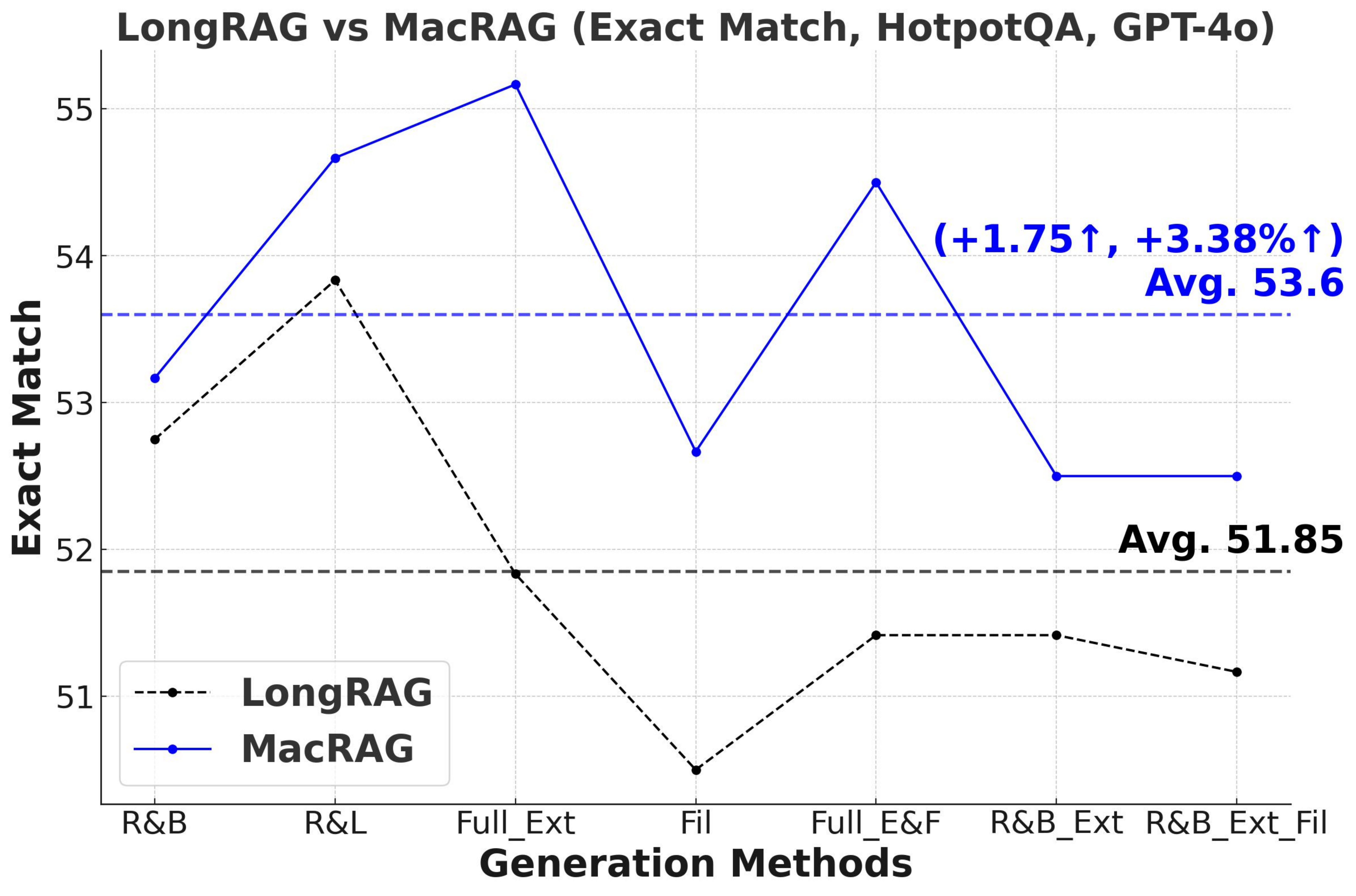}
\includegraphics[width=0.24\textwidth]{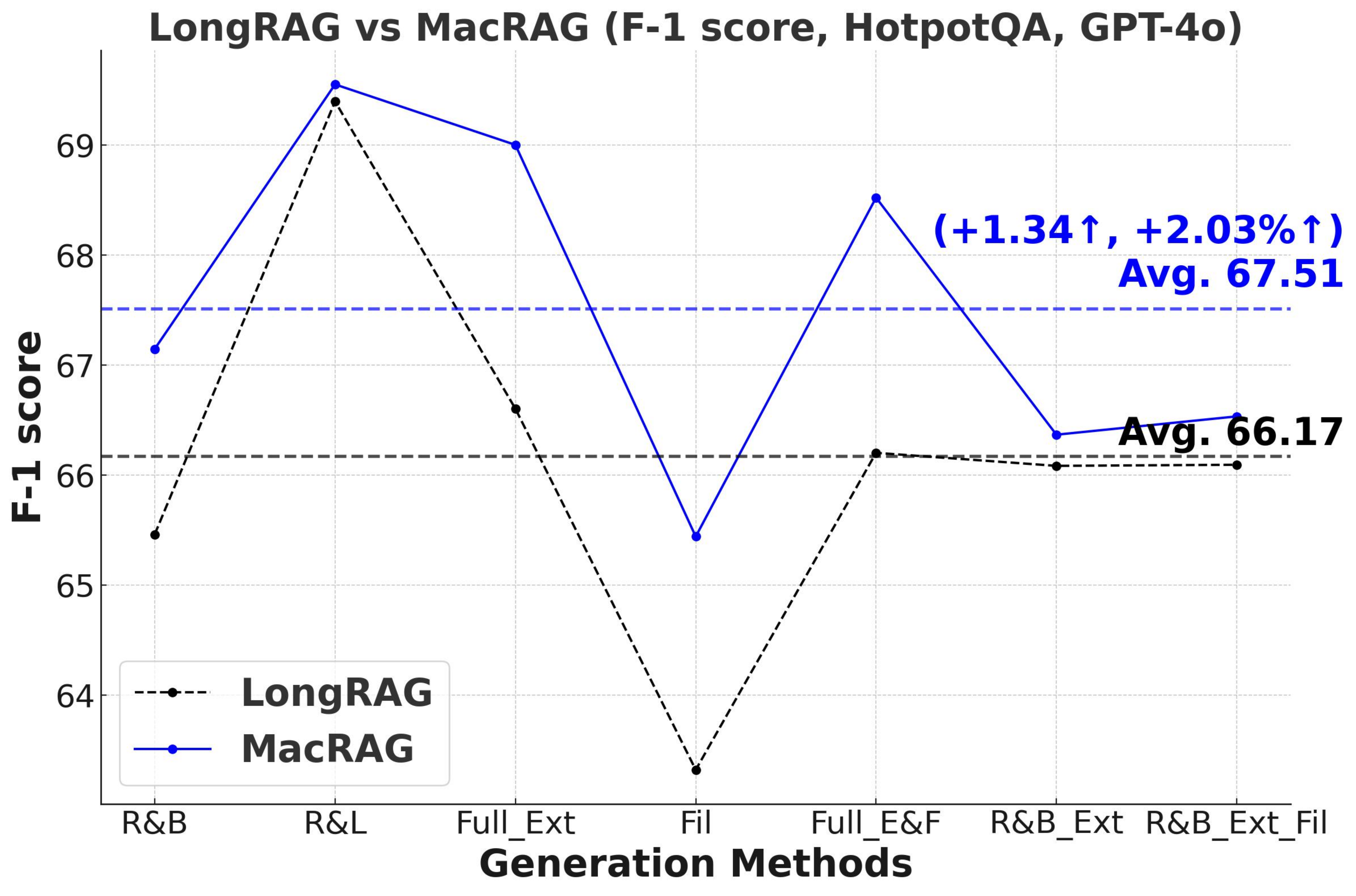}
\includegraphics[width=0.24\textwidth]{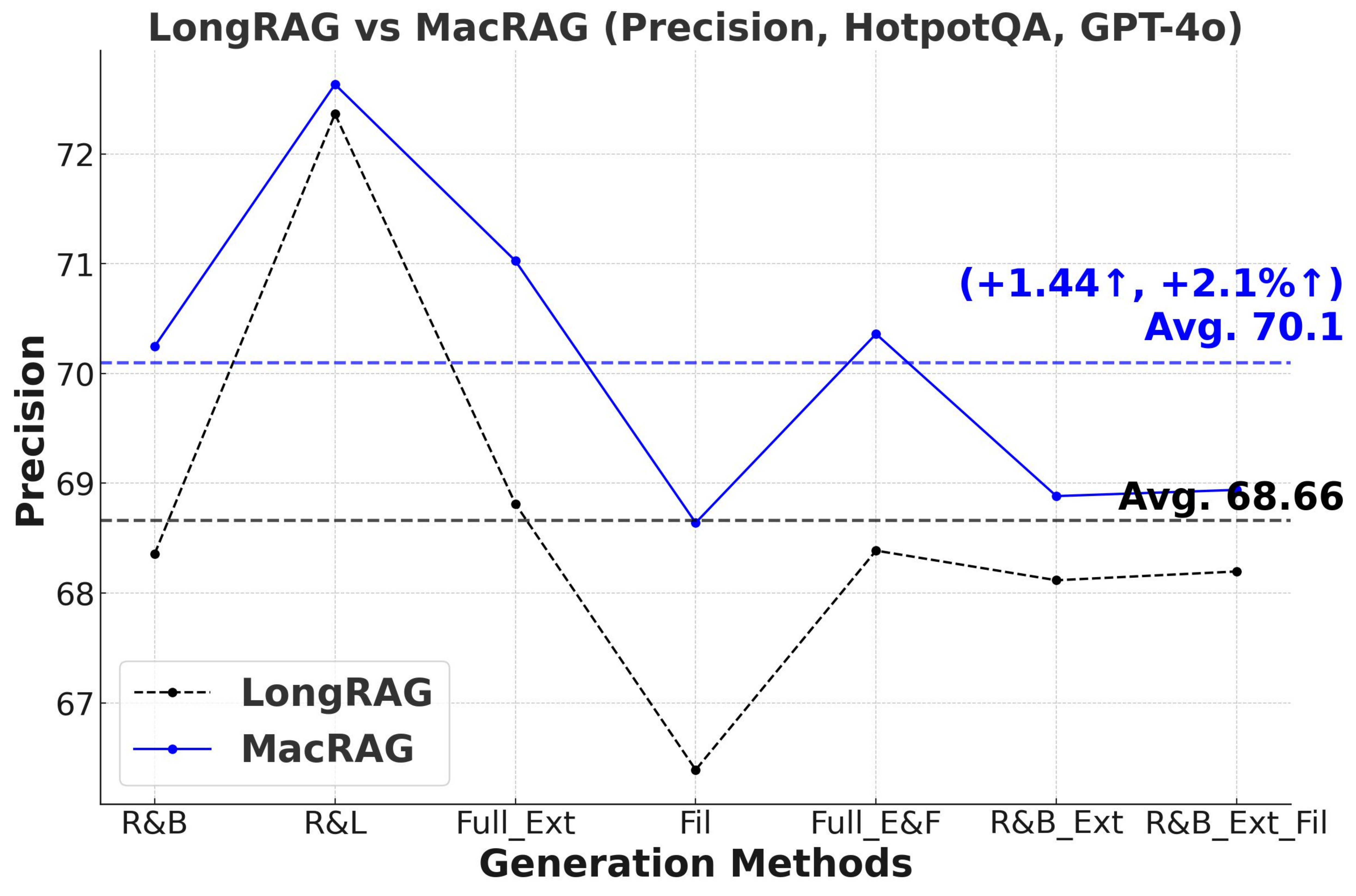}
\includegraphics[width=0.24\textwidth]{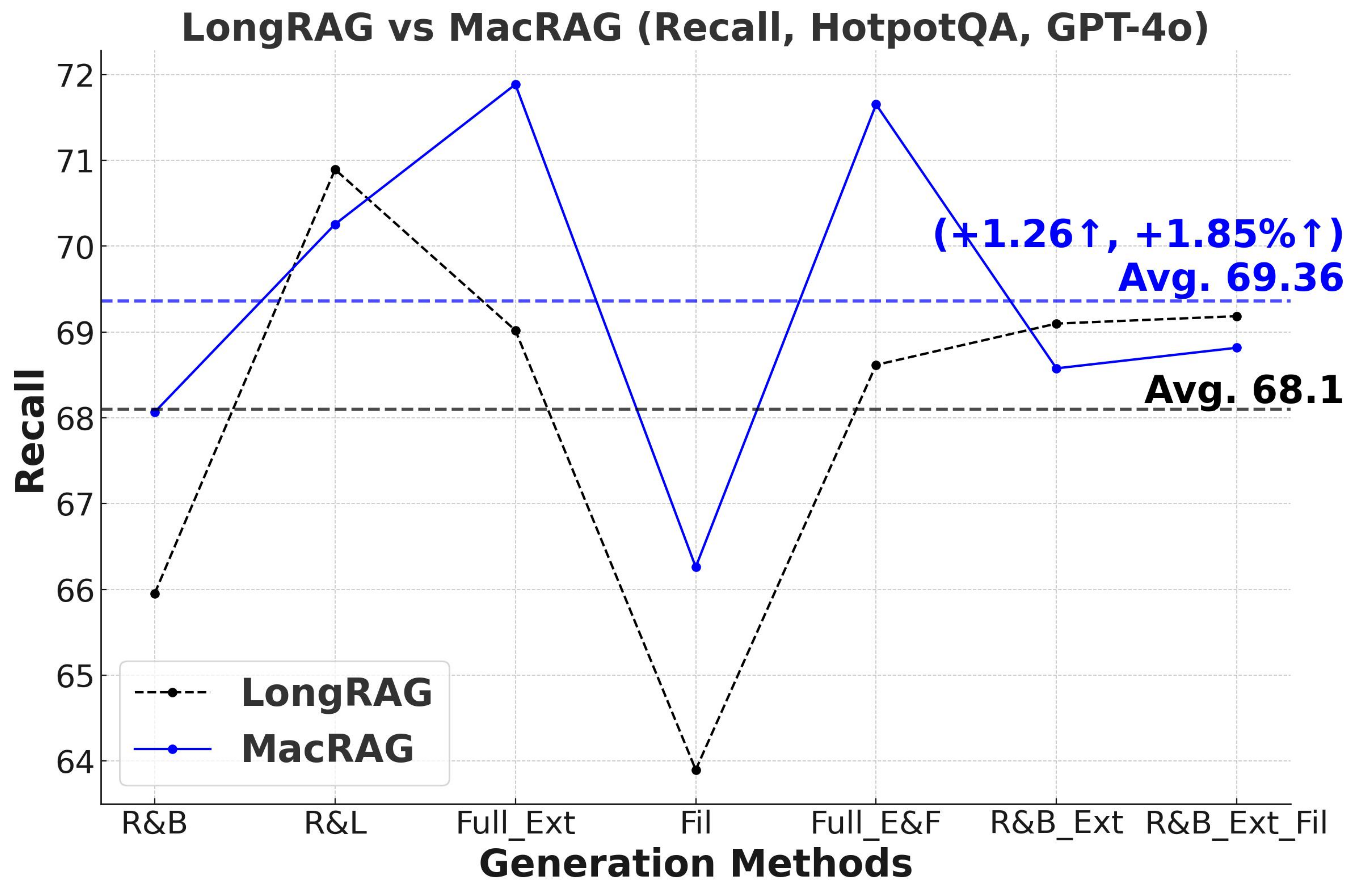}

\includegraphics[width=0.24\textwidth]{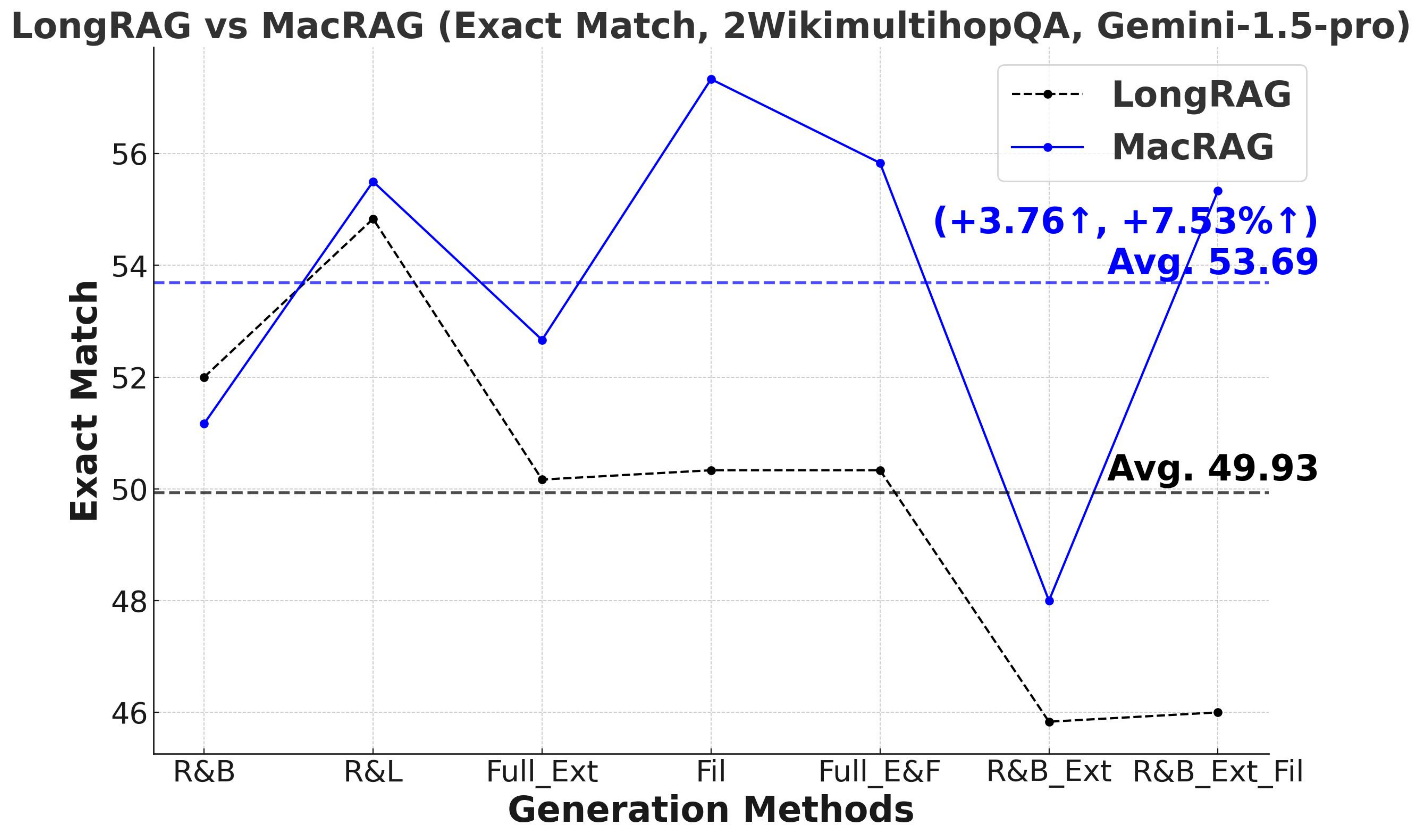}
\includegraphics[width=0.24\textwidth]{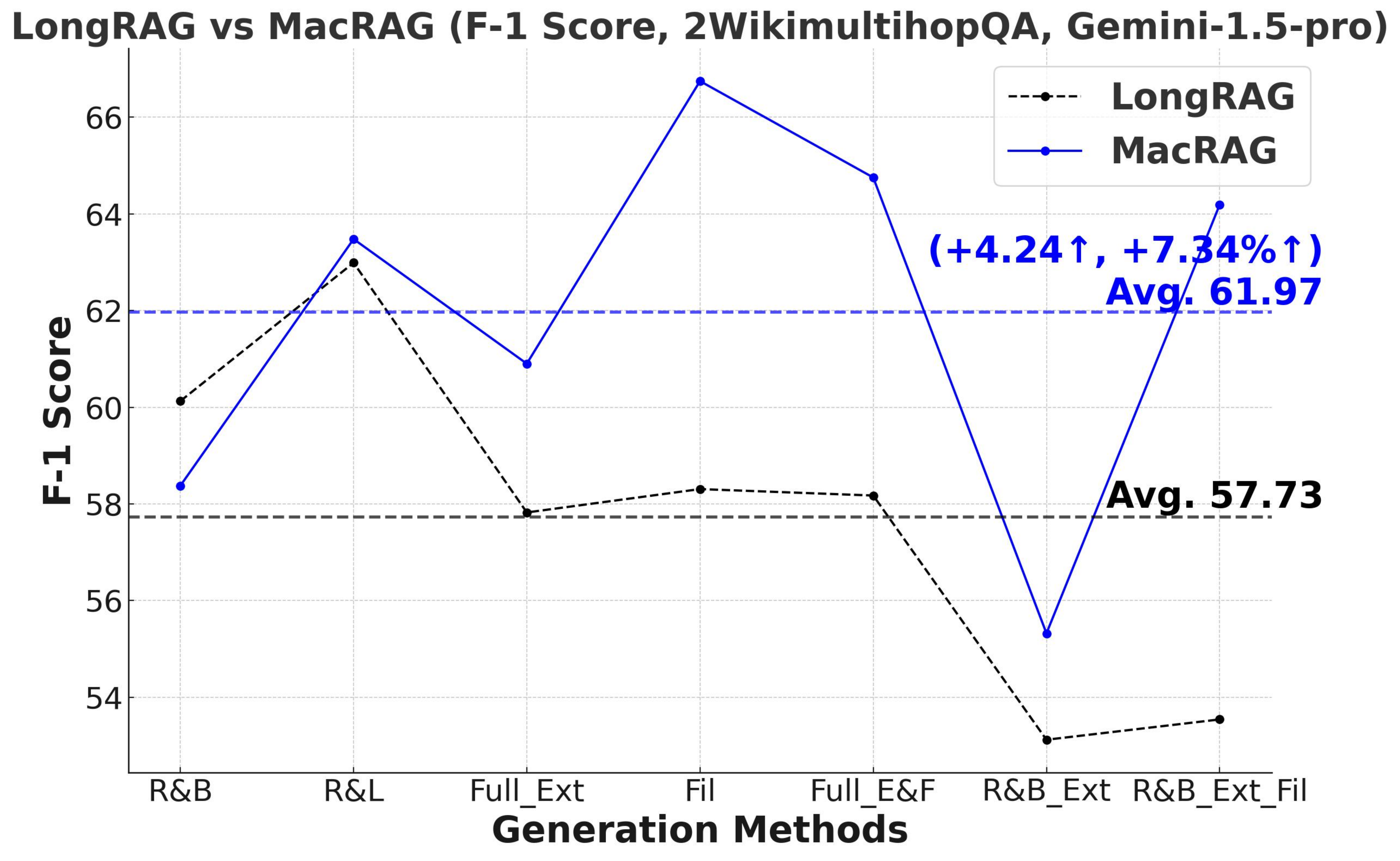}
\includegraphics[width=0.24\textwidth]{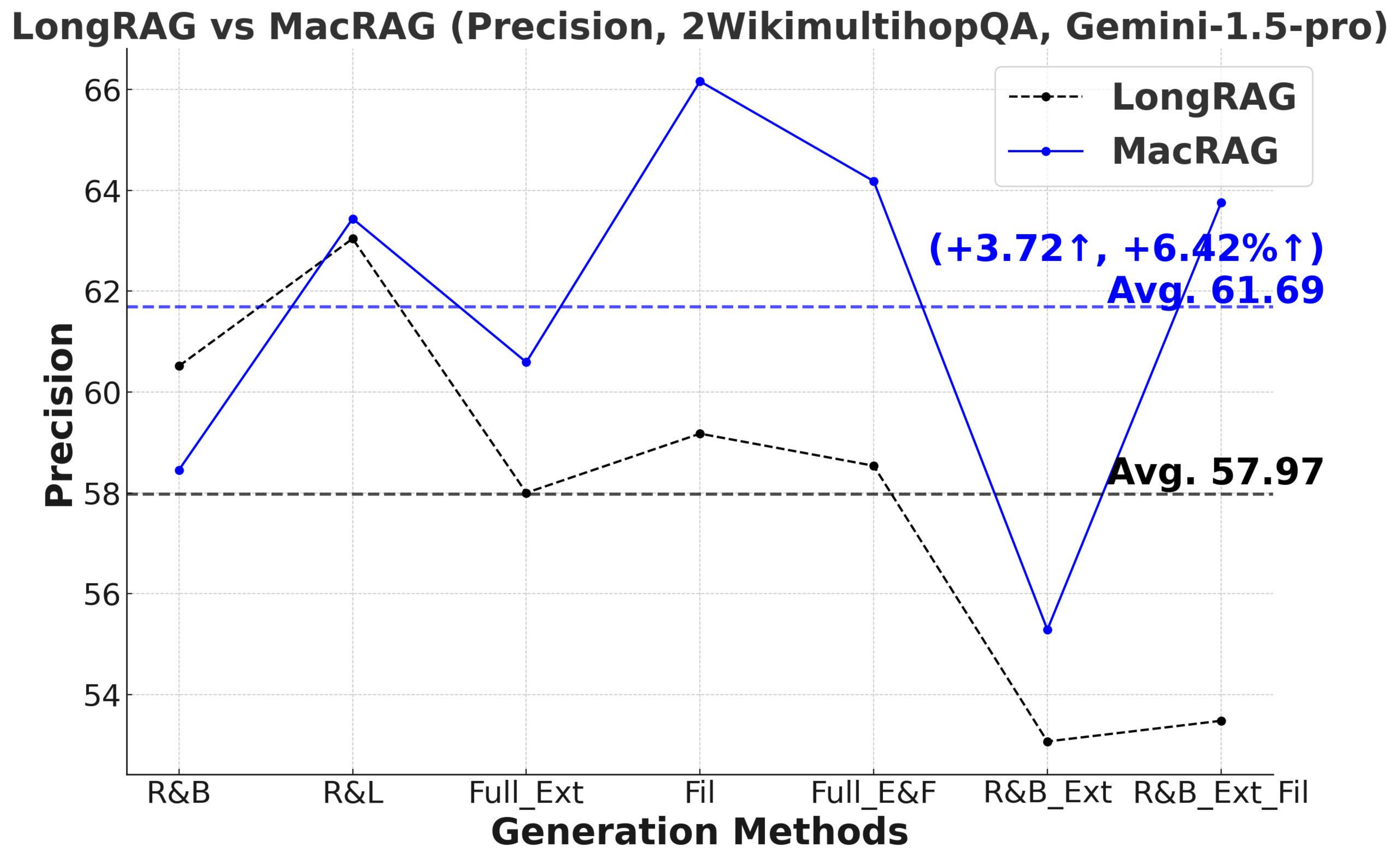}
\includegraphics[width=0.24\textwidth]{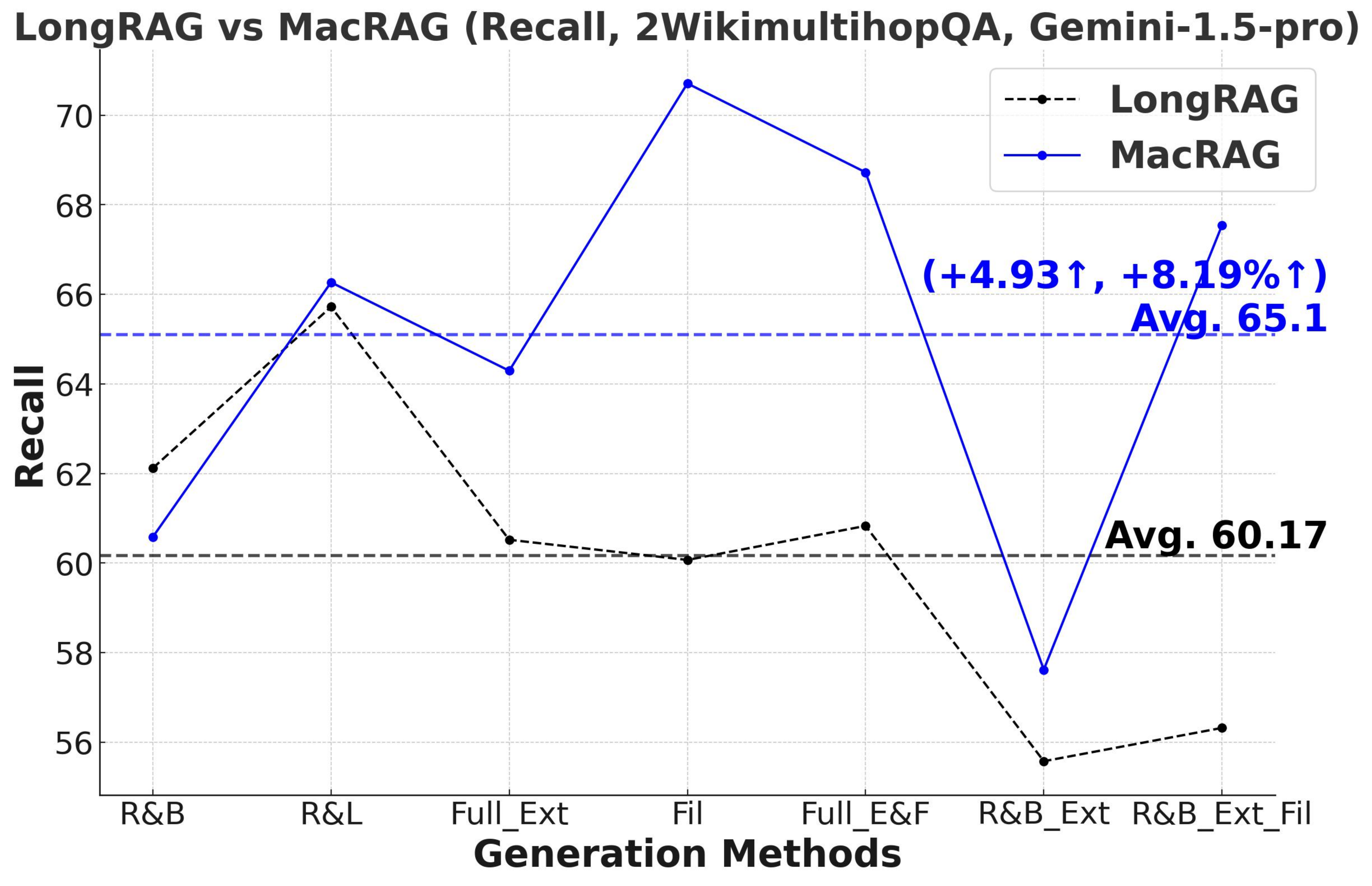}

\includegraphics[width=0.24\textwidth]{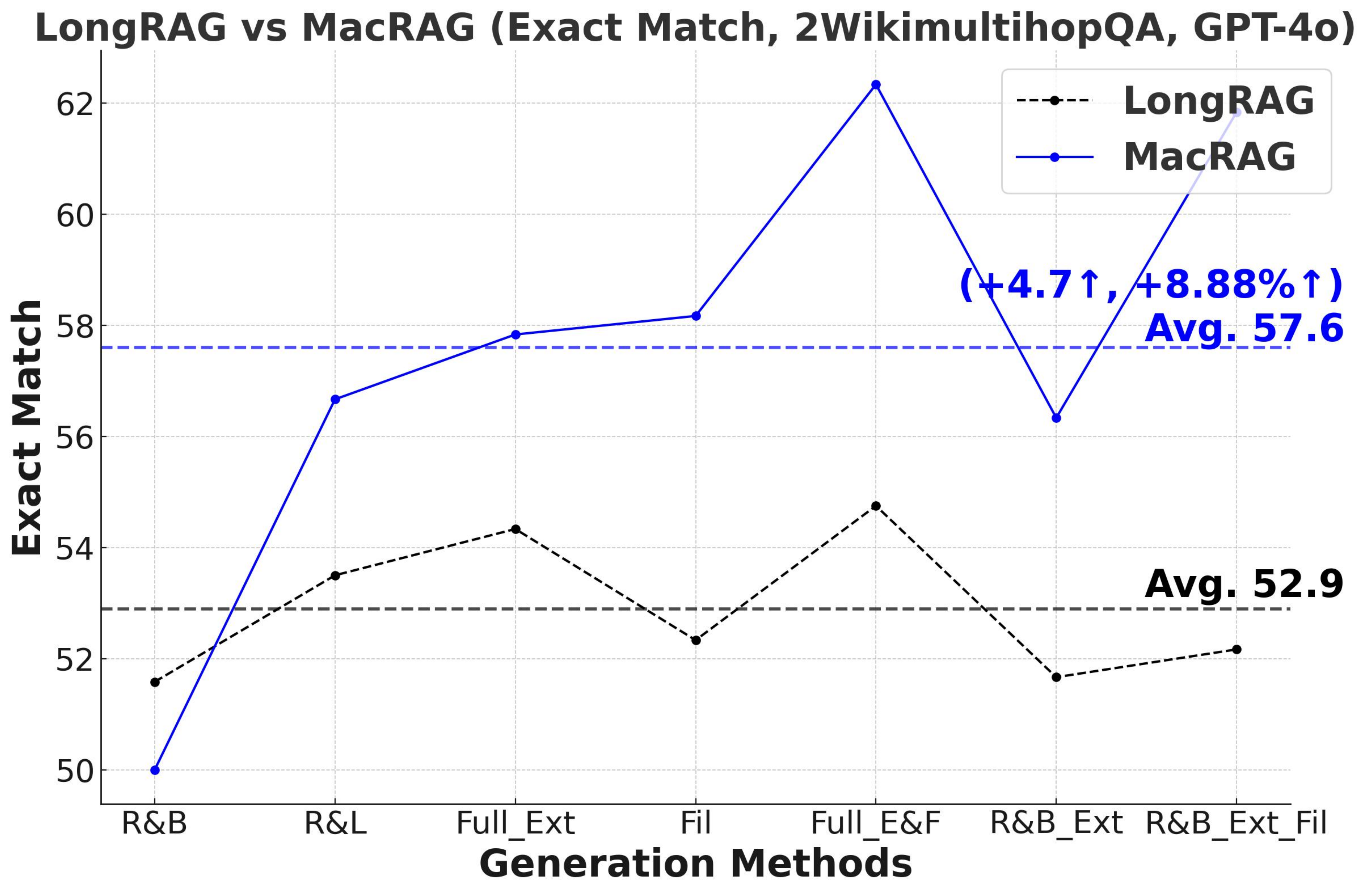}
\includegraphics[width=0.24\textwidth]{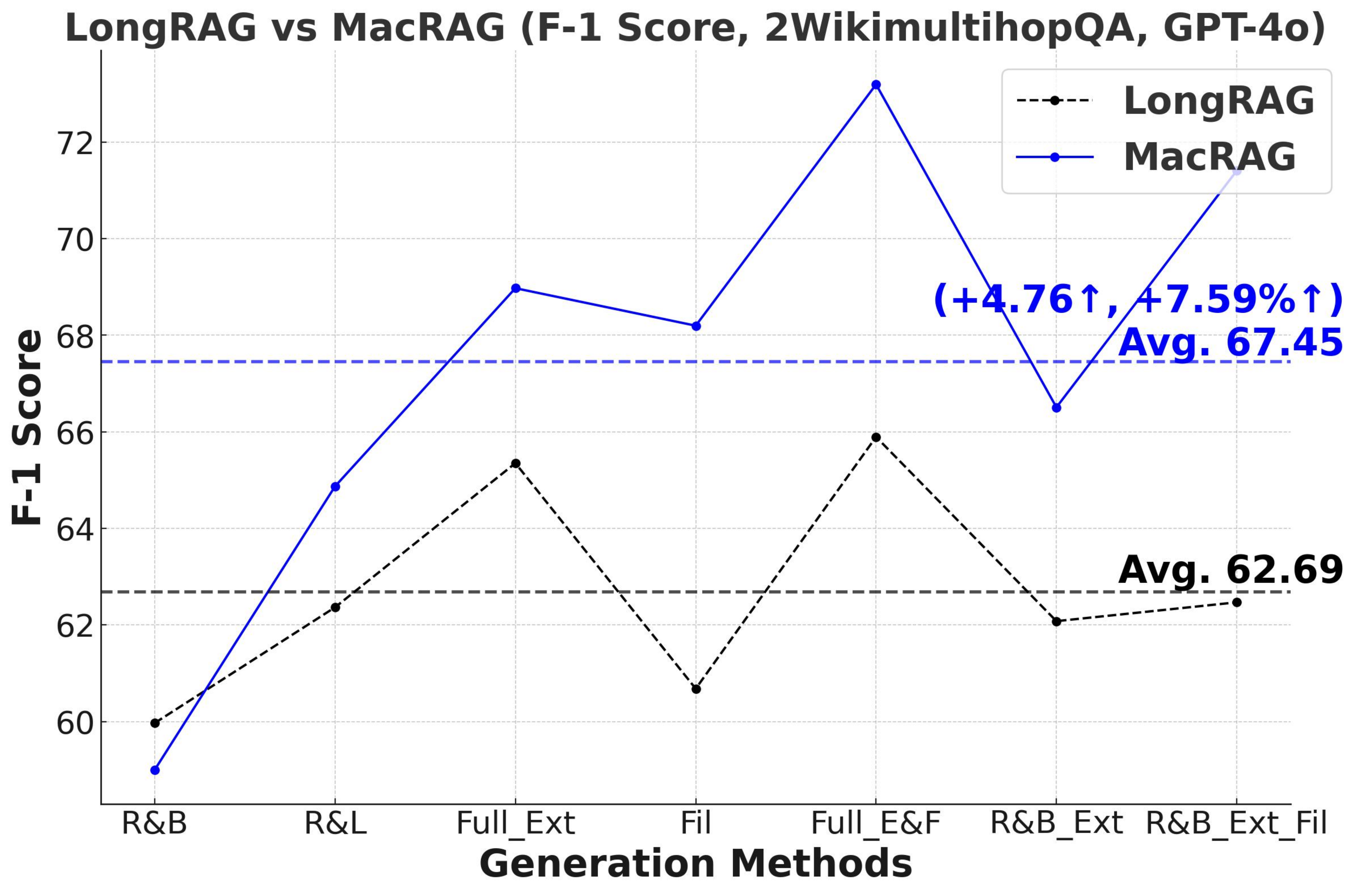}
\includegraphics[width=0.24\textwidth]{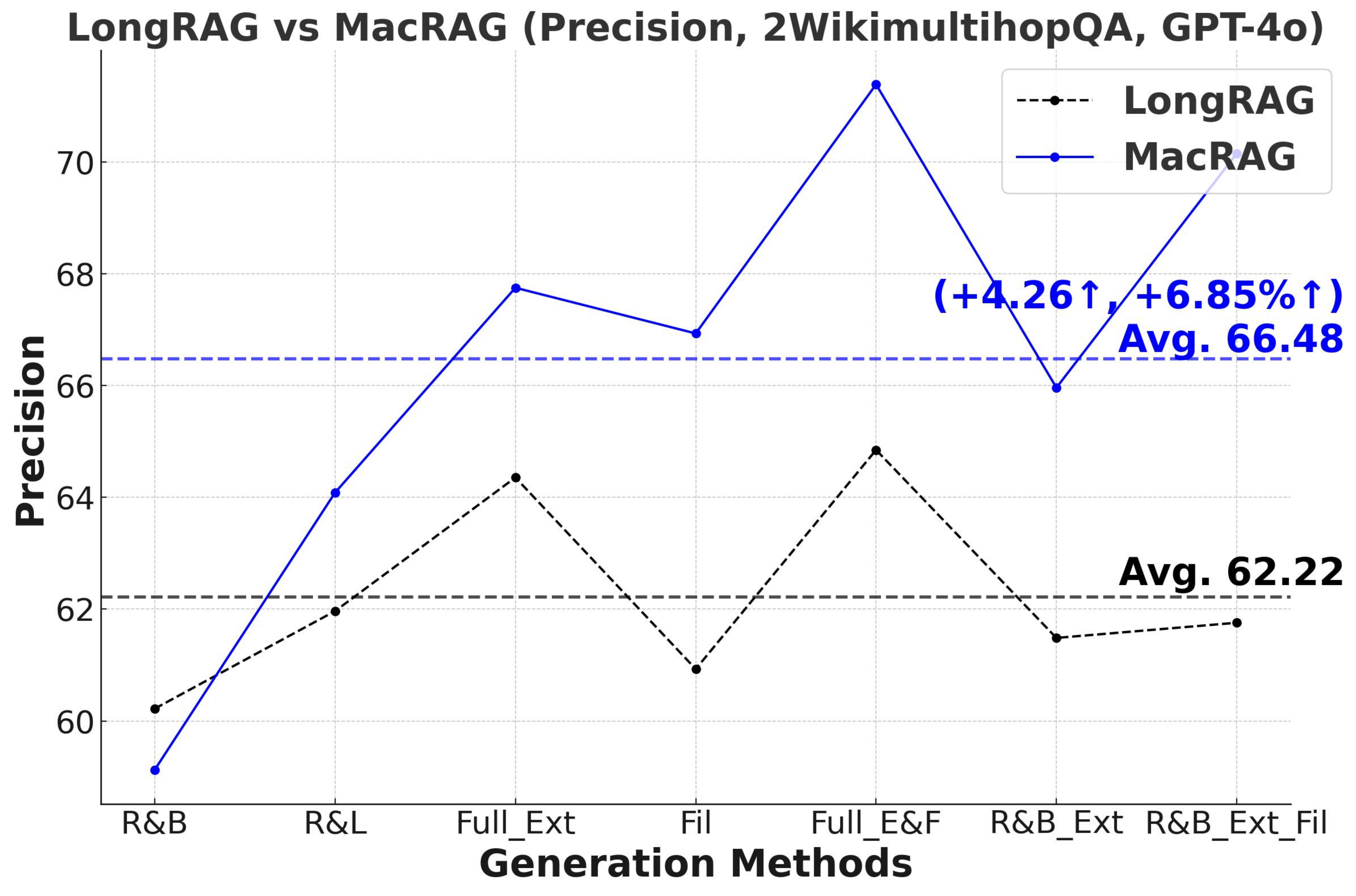}
\includegraphics[width=0.24\textwidth]{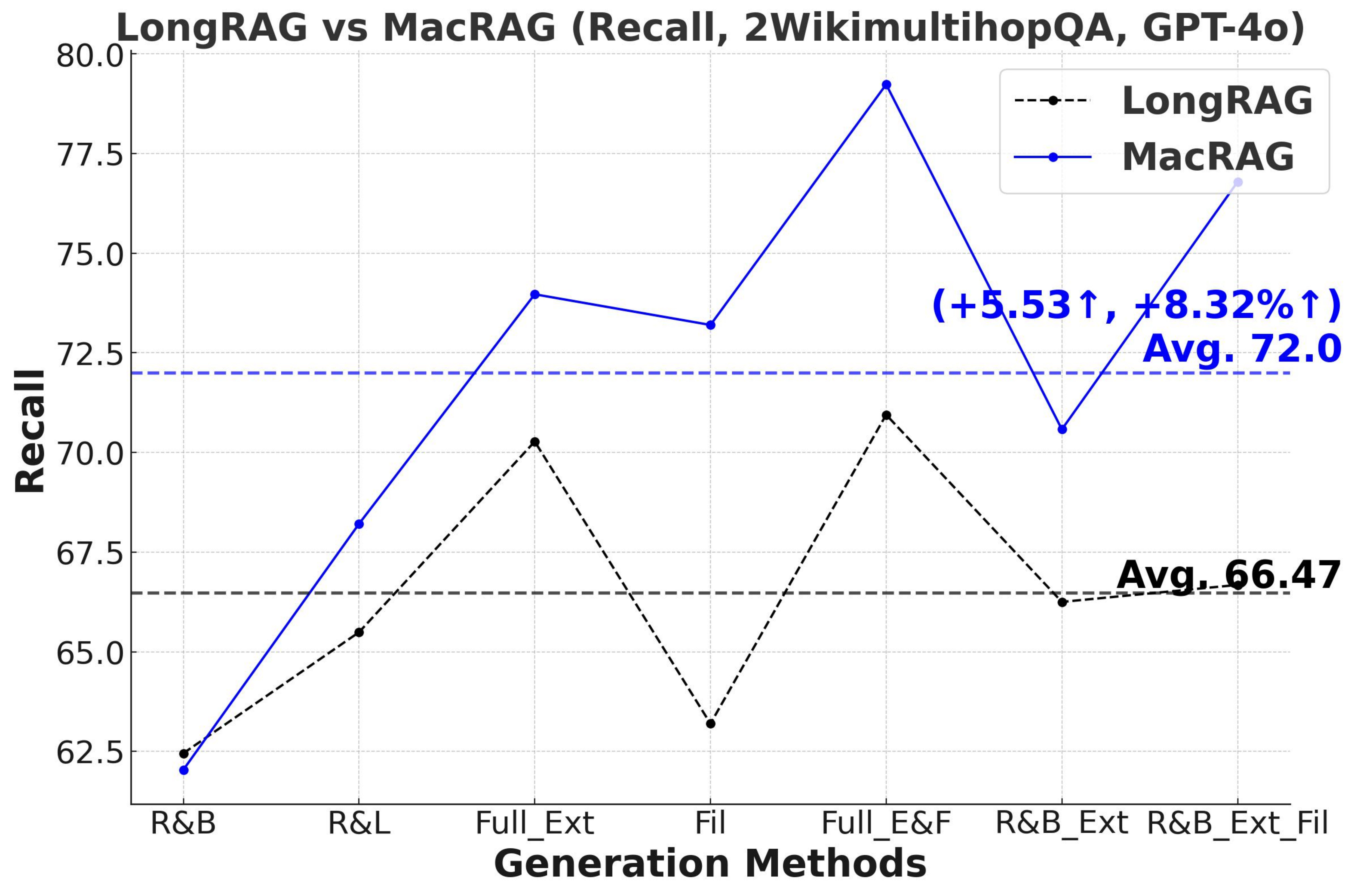}

\includegraphics[width=0.24\textwidth]{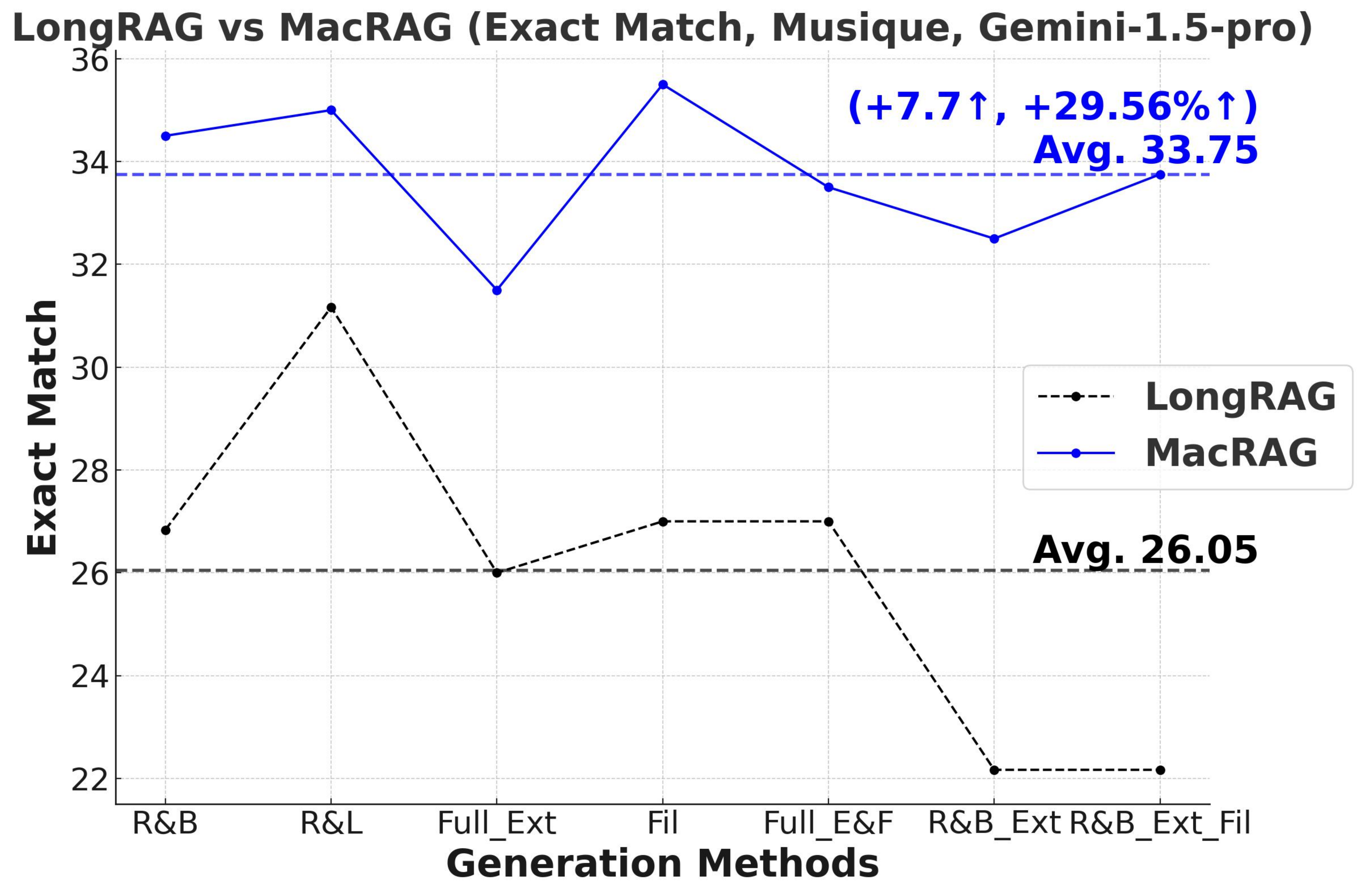}
\includegraphics[width=0.24\textwidth]{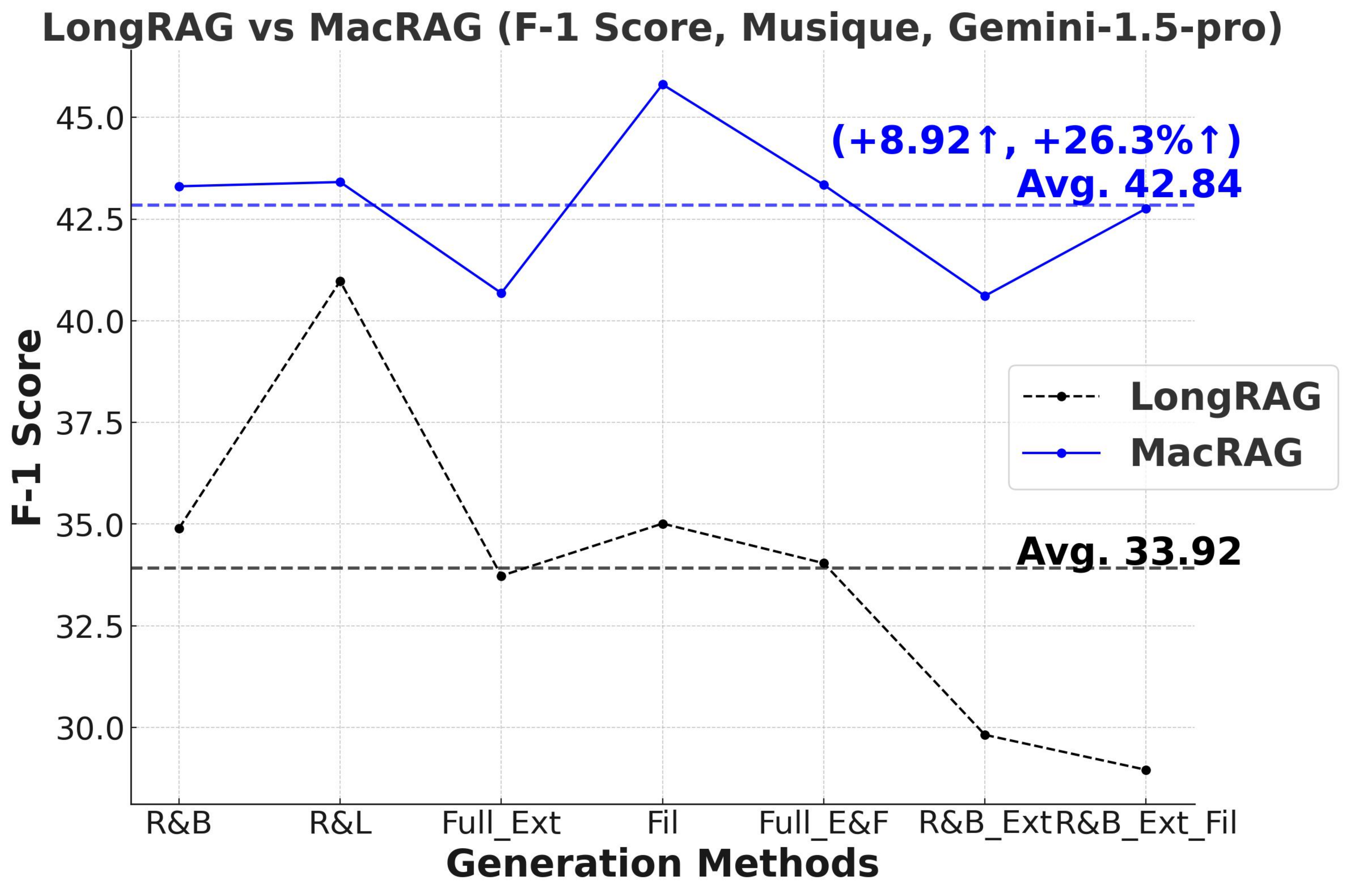}
\includegraphics[width=0.24\textwidth]{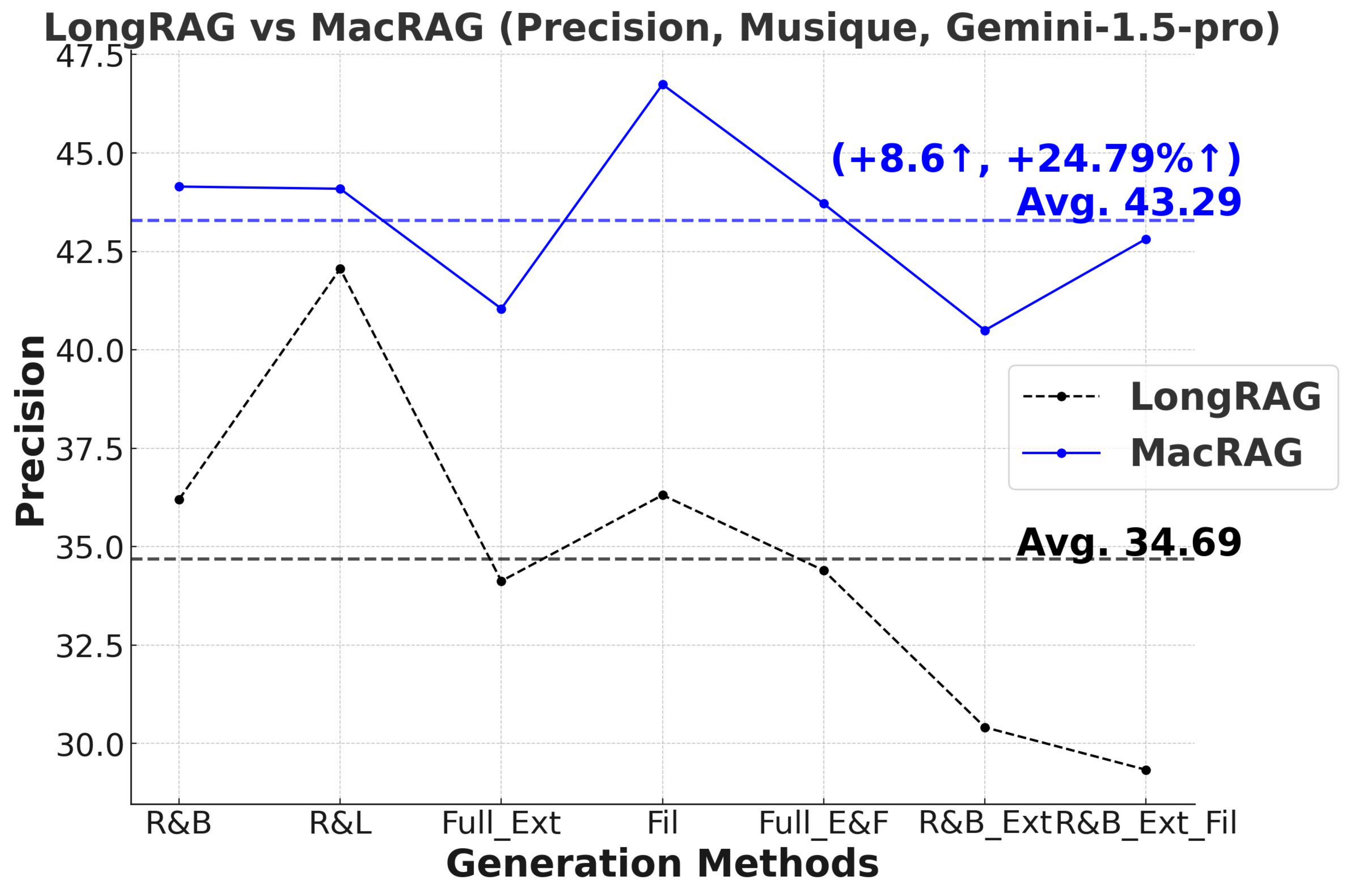}
\includegraphics[width=0.24\textwidth]{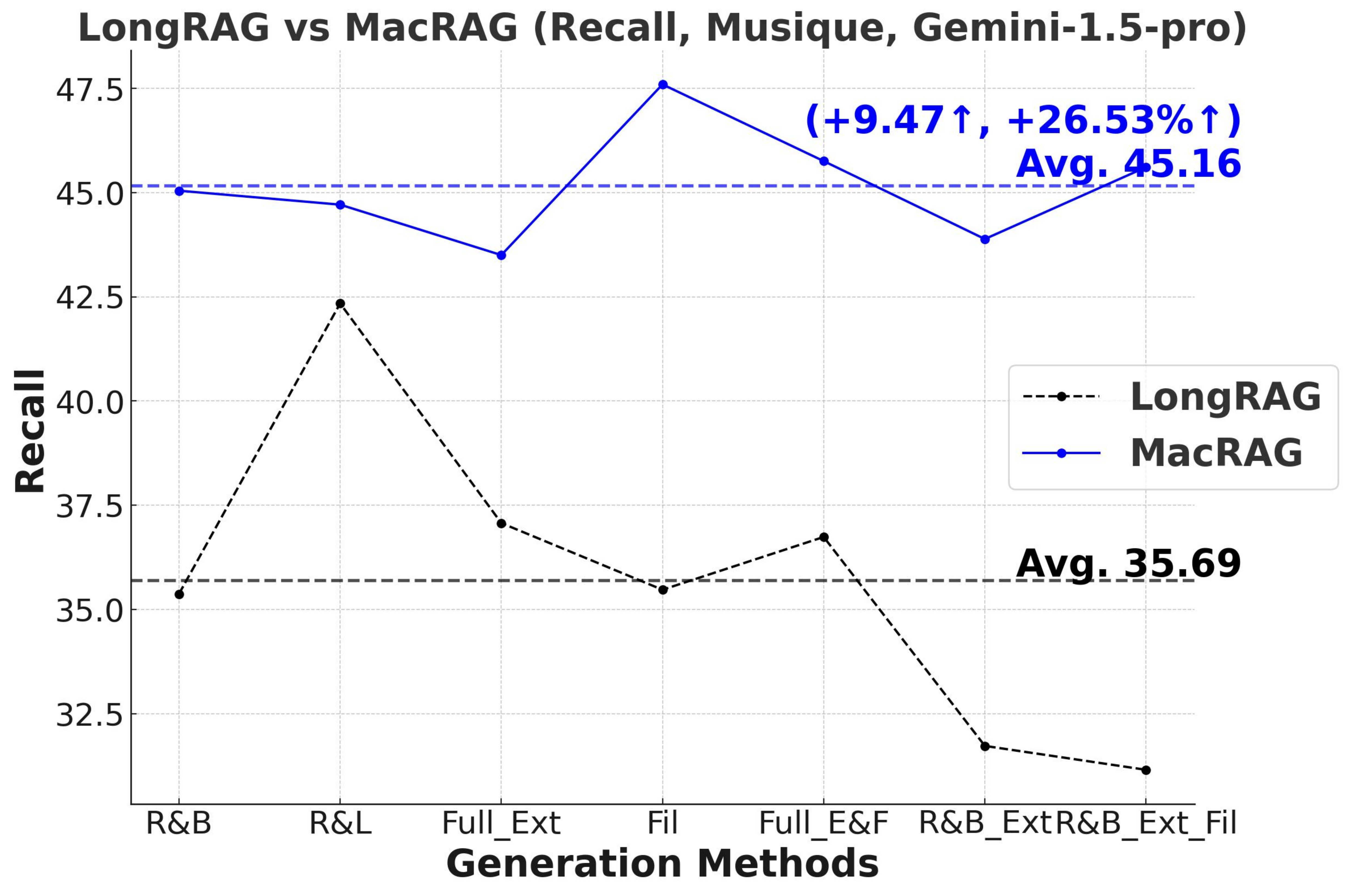}

\includegraphics[width=0.24\textwidth]{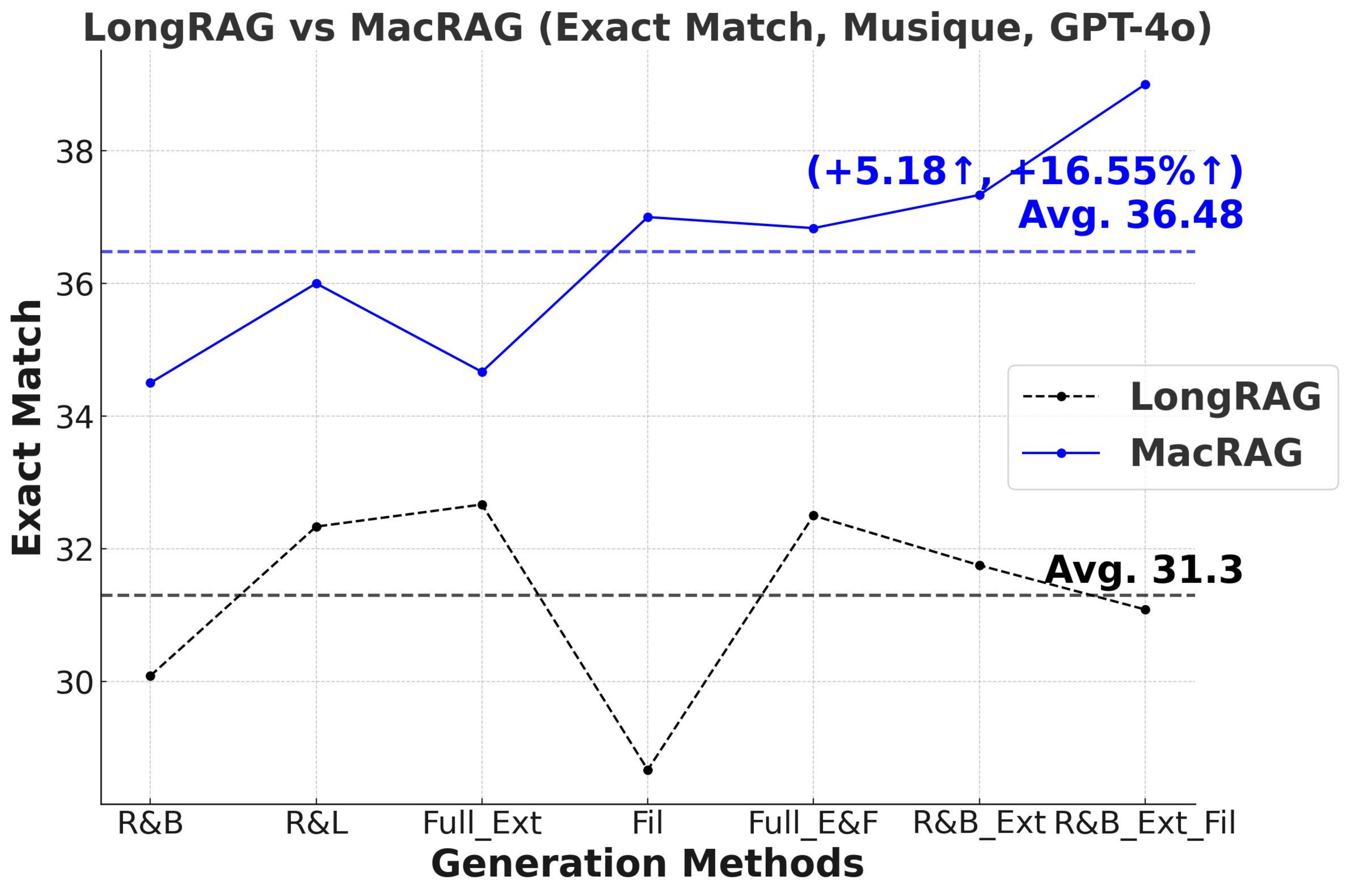}
\includegraphics[width=0.24\textwidth]{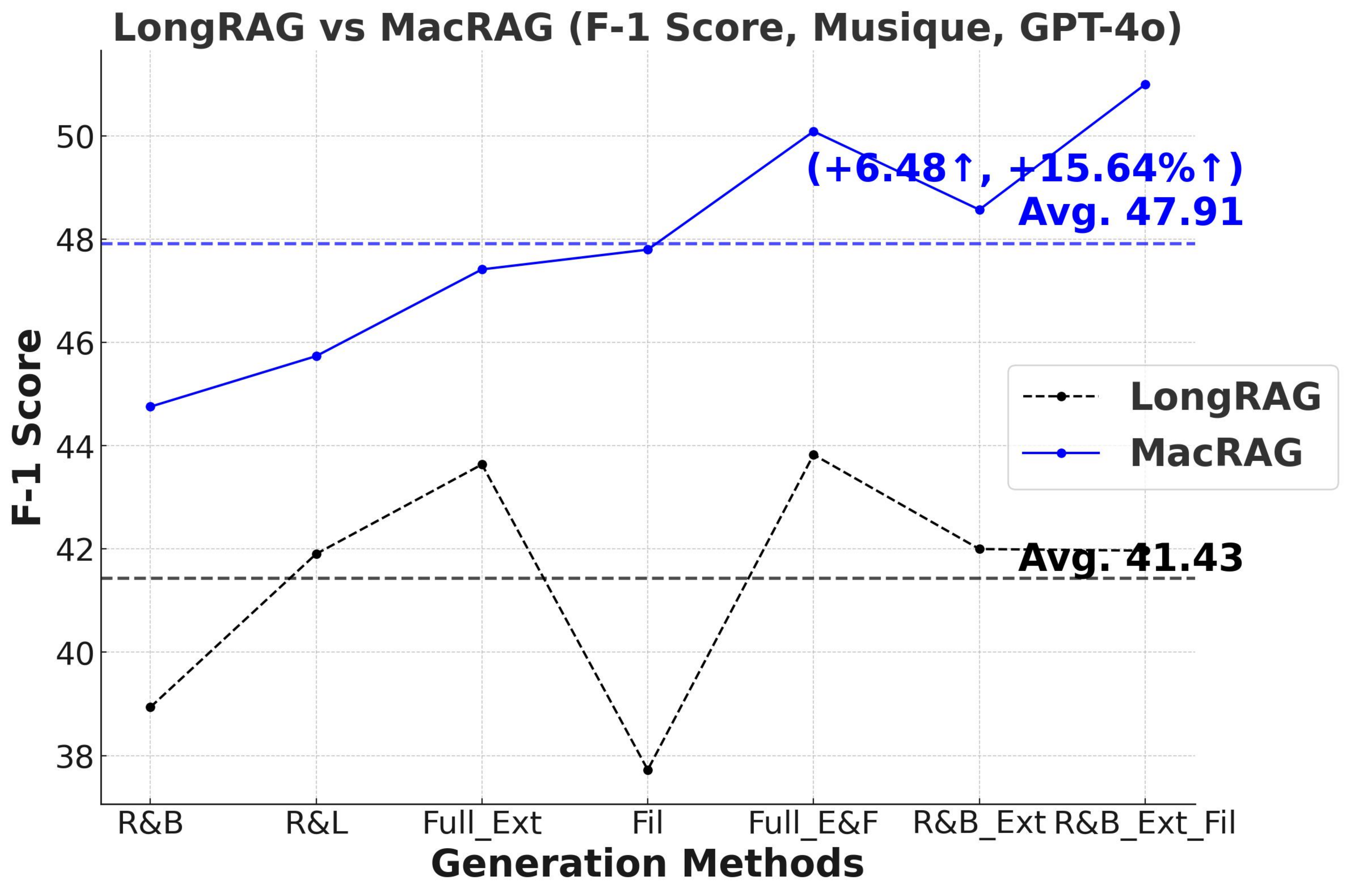}
\includegraphics[width=0.24\textwidth]{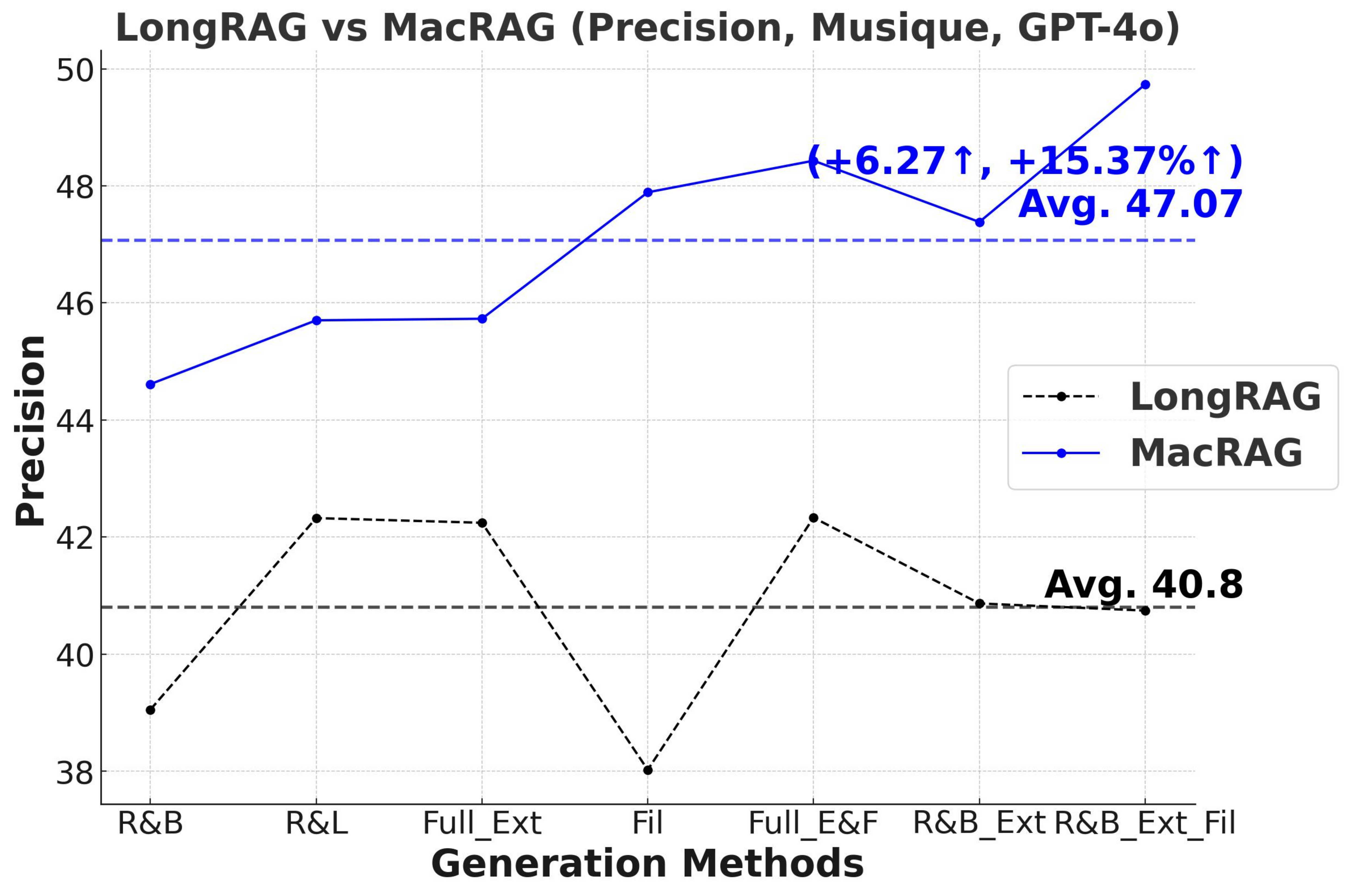}
\includegraphics[width=0.24\textwidth]{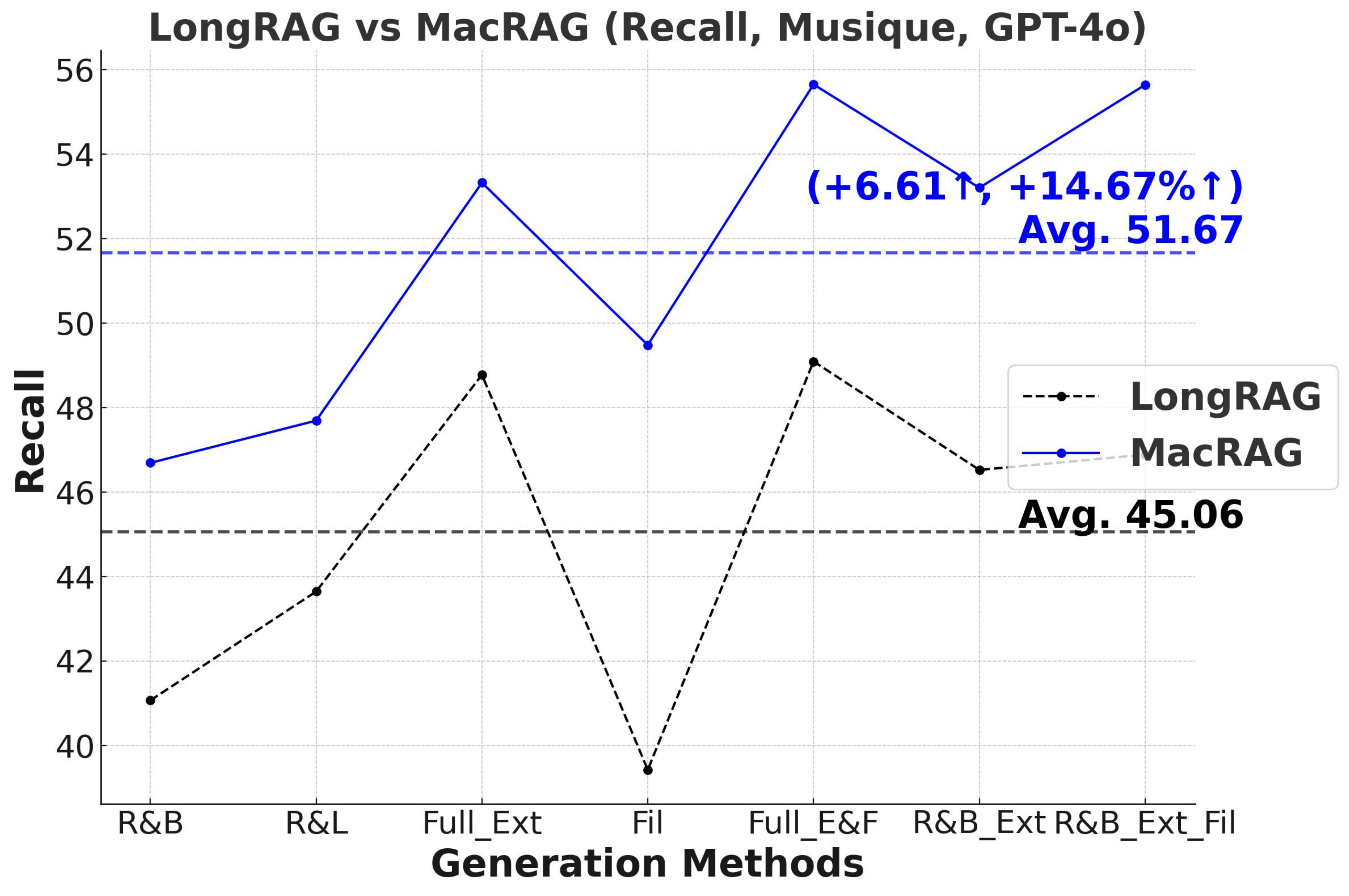}

\caption{Performances of LongRAG and MacRAG regarding the fours metrics (Exact Match, F1-score, Precision, Recall) for three datasets (HotpotQA, 2WikimultihopQA, and Musique) and two LLMs (Gemini-1.5-pro and GPT-4o). Each row corresponds to a combination of dataset and LLM, and each column represents one of the metrics.}
\label{fig_appendix_comparison_overall_metics_all_datasets_for_Gemini_GPT}
\end{figure*}

\begin{table*}[t!]
\centering
\scriptsize
\setlength{\tabcolsep}{0.7pt}
\begin{tabular}{|c|c|c|c|c|c|c|c|c|}
\hline
\multicolumn{9}{|c|}{\textbf{HotpotQA}} \\
\hline
\textbf{Method ($k_2=7$)} & \textbf{R\&B} & \textbf{R\&L} & \textbf{Full\_Ext} & \textbf{Fil} & \textbf{Full\_E\&F} & \textbf{R\&B\_Ext} & \textbf{R\&B\_Ext\_Fil} & \textbf{Average} \\
\hline
LongRAG & 63.59 & 63.93 & 61.93 & 63.38 & 61.95 & 57.86 & 57.53 & 61.45 \\
\textbf{MacRAG}
 & 63.02 (\textbf{-0.57$\downarrow$})
 & 66.76 (\textbf{+2.83$\uparrow$})
 & 64.90 (\textbf{+2.97$\uparrow$})
 & 65.96 (\textbf{+2.58$\uparrow$})
 & 64.40 (\textbf{+2.45$\uparrow$})
 & 60.59 (\textbf{+2.73$\uparrow$})
 & 62.14 (\textbf{+4.61$\uparrow$})
 & \textbf{63.97} (\textbf{+2.52$\uparrow$}, \textbf{+4.10\%$\uparrow$}) \\
 
\hline

\multicolumn{9}{|c|}{\textbf{2WikimultihopQA}} \\
\hline
\textbf{Method ($k_2=7$)} & \textbf{R\&B} & \textbf{R\&L} & \textbf{Full\_Ext} & \textbf{Fil} & \textbf{Full\_E\&F} & \textbf{R\&B\_Ext} & \textbf{R\&B\_Ext\_Fil} & \textbf{Average} \\
\hline
LongRAG & 60.13 & 62.99 & 57.82 & 58.31 & 58.17 & 53.12 & 53.54 & 57.73 \\
\textbf{MacRAG}
 & 58.38 (\textbf{-1.75$\downarrow$})
 & 63.48 (\textbf{+0.49$\uparrow$})
 & 60.90 (\textbf{+3.08$\uparrow$})
 & 66.74 (\textbf{+8.43$\uparrow$})
 & 64.75 (\textbf{+6.58$\uparrow$})
 & 55.32 (\textbf{+2.20$\uparrow$})
 & 64.19 (\textbf{+10.65$\uparrow$})
 & \textbf{61.85} (\textbf{+4.12$\uparrow$}, \textbf{+7.14\%$\uparrow$}) \\
\hline

\multicolumn{9}{|c|}{\textbf{Musique}} \\
\hline
\textbf{Method ($k_2=7$)} & \textbf{R\&B} & \textbf{R\&L} & \textbf{Full\_Ext} & \textbf{Fil} & \textbf{Full\_E\&F} & \textbf{R\&B\_Ext} & \textbf{R\&B\_Ext\_Fil} & \textbf{Average} \\
\hline
LongRAG & 34.90 & 40.97 & 33.73 & 35.01 & 34.04 & 29.82 & 28.96 & 33.92 \\
\textbf{MacRAG}
 & 43.31 (\textbf{+8.41$\uparrow$})
 & 43.41 (\textbf{+2.44$\uparrow$})
 & 40.68 (\textbf{+6.96$\uparrow$})
 & 45.81 (\textbf{+10.80$\uparrow$})
 & 43.34 (\textbf{+9.30$\uparrow$})
 & 40.61 (\textbf{+10.79$\uparrow$})
 & 42.76 (\textbf{+13.80$\uparrow$})
 & \textbf{42.84} (\textbf{+8.92$\uparrow$}), \textbf{+26.30\%$\uparrow$}) \\
\hline

\end{tabular}
\caption{Experimental results of Gemini-1.5-pro with F1-score for comparing LongRAG vs MacRAG across HotpotQA, 2WikimultihopQA, and Musique datasets. 
The experiments conducted with the same hyper-parameter (k1=100 , k2=7) which is the reported best parameter of LongRAG \cite{zhao2024longrag}.
The columns represent various evaluation settings: R\&B (Retrieval and Base), R\&L (Retrieval and Long), Full\_Ext (Extraction from Full Document), Fil (Filtering), Full\_E\&F (Extraction and Filtering combined), R\&B\_Ext (Extraction from Top-$k_2$ Chunks), and R\&B\_Ext\_Fil (Extraction and Filtering from Top-$k_2$ Chunks). The absolute gains and relative percentage improvements from applying MacRAG to LongRAG are displayed in parentheses. }
\label{tbl_appendix_experimental_results_MacRAG_vs_LongRAG_gemini_detail}
\end{table*}
\begin{table*}[t!]
\centering
\scriptsize
\setlength{\tabcolsep}{1.5pt}
\begin{tabular}{|c|c|c|c|c|c|c|c|c|}
\hline
\multicolumn{9}{|c|}{\textbf{HotpotQA}} \\
\hline
\textbf{$k_2=7$, marco-miniLM} & \textbf{R\&B} & \textbf{R\&L} & \textbf{Full\_Ext} & \textbf{Fil} & \textbf{Full\_E\&F} & \textbf{R\&B\_Ext} & \textbf{R\&B\_Ext\_Fil} & \textbf{Average} \\
\hline
LongRAG & 65.46 & 69.40 & 66.60 & 63.32 & 66.20 & 66.08 & 66.09 & 66.16 \\
\textbf{MacRAG}
 & 67.15 (\textbf{+1.69$\uparrow$})
 & 69.55 (\textbf{+0.15$\uparrow$})
 & 69.00 (\textbf{+2.40$\uparrow$})
 & 65.44 (\textbf{+2.12$\uparrow$})
 & 68.52 (\textbf{+2.32$\uparrow$})
 & 66.36 (\textbf{+0.28$\uparrow$})
 & 66.53 (\textbf{+0.44$\uparrow$})
 & \textbf{67.22} (\textbf{+1.06$\uparrow$}) \\
\hline

\textbf{$k_2=12$, marco-miniLM} & \textbf{R\&B} & \textbf{R\&L} & \textbf{Full\_Ext} & \textbf{Fil} & \textbf{Full\_E\&F} & \textbf{R\&B\_Ext} & \textbf{R\&B\_Ext\_Fil} & \textbf{Average} \\
\hline
LongRAG & 67.97 & 69.80 & 67.91 & 63.56 & 67.82 & 66.91 & 66.44 & 67.20 \\
\textbf{MacRAG}
 & 68.41 (\textbf{+0.44$\uparrow$})
 & 69.64 (\textbf{-0.16$\downarrow$})
 & 69.95 (\textbf{+2.04$\uparrow$})
 & 65.42 (\textbf{+1.86$\uparrow$})
 & 69.09 (\textbf{+1.27$\uparrow$})
 & 67.57 (\textbf{+0.66$\uparrow$})
 & 66.95 (\textbf{+0.51$\uparrow$})
 & \textbf{68.00} (\textbf{+0.80$\uparrow$}) \\
\hline

\textbf{$k_2=7$, bge-m3} & \textbf{R\&B} & \textbf{R\&L} & \textbf{Full\_Ext} & \textbf{Fil} & \textbf{Full\_E\&F} & \textbf{R\&B\_Ext} & \textbf{R\&B\_Ext\_Fil} & \textbf{Average} \\
\hline
LongRAG & 67.67 & 67.99 & 68.96 & 64.30 & 68.49 & 66.73 & 66.10 & 67.18 \\
\textbf{MacRAG}
 & 67.59 (\textbf{-0.08$\downarrow$})
 & 68.63 (\textbf{+0.64$\uparrow$})
 & 70.53 (\textbf{+1.57$\uparrow$})
 & 65.56 (\textbf{+1.26$\uparrow$})
 & 70.29 (\textbf{+1.80$\uparrow$})
 & 66.59 (\textbf{-0.14$\downarrow$})
 & 67.32 (\textbf{+1.22$\uparrow$})
 & \textbf{68.07} (\textbf{+0.89$\uparrow$}) \\
\hline

\textbf{$k_2=12$, bge-m3} & \textbf{R\&B} & \textbf{R\&L} & \textbf{Full\_Ext} & \textbf{Fil} & \textbf{Full\_E\&F} & \textbf{R\&B\_Ext} & \textbf{R\&B\_Ext\_Fil} & \textbf{Average} \\
\hline
LongRAG & 68.57 & 67.65 & 70.31 & 64.74 & 70.14 & 67.63 & 67.49 & 68.08 \\
\textbf{MacRAG}
 & 67.88 (\textbf{-0.69$\downarrow$})
 & 69.87 (\textbf{+2.22$\downarrow$})
 & 70.72 (\textbf{+0.41$\uparrow$})
 & 65.98 (\textbf{+1.24$\uparrow$})
 & 70.66 (\textbf{+0.52$\uparrow$})
 & 69.05 (\textbf{+1.42$\uparrow$})
 & 67.84 (\textbf{+0.35$\uparrow$})
 & \textbf{68.86} (\textbf{+0.78$\uparrow$}) \\
\hline

\multicolumn{9}{|c|}{\textbf{2WikimultihopQA}} \\
\hline
\textbf{$k_2=7$, marco-miniLM} & \textbf{R\&B} & \textbf{R\&L} & \textbf{Full\_Ext} & \textbf{Fil} & \textbf{Full\_E\&F} & \textbf{R\&B\_Ext} & \textbf{R\&B\_Ext\_Fil} & \textbf{Average} \\
\hline
LongRAG & 59.97 & 62.37 & 65.35 & 60.68 & 65.89 & 62.08 & 62.47 & 62.69 \\
\textbf{MacRAG}
 & 59.00 (\textbf{-0.97$\downarrow$})
 & 64.87 (\textbf{+2.50$\uparrow$})
 & 68.97 (\textbf{+3.62$\uparrow$})
 & 68.20 (\textbf{+7.52$\uparrow$})
 & 73.19 (\textbf{+7.30$\uparrow$})
 & 66.50 (\textbf{+4.42$\uparrow$})
 & 71.40 (\textbf{+8.93$\uparrow$})
 & \textbf{67.45} (\textbf{+4.76$\uparrow$}) \\
\hline

\textbf{$k_2=12$, marco-miniLM} & \textbf{R\&B} & \textbf{R\&L} & \textbf{Full\_Ext} & \textbf{Fil} & \textbf{Full\_E\&F} & \textbf{R\&B\_Ext} & \textbf{R\&B\_Ext\_Fil} & \textbf{Average} \\
\hline
LongRAG & 62.64 & 65.43 & 70.67 & 62.06 & 70.66 & 67.56 & 66.60 & 66.52 \\
\textbf{MacRAG}
 & 63.47 (\textbf{+0.83$\uparrow$})
 & 67.06 (\textbf{+1.63$\uparrow$})
 & 70.61 (\textbf{-0.06$\downarrow$})
 & 67.84 (\textbf{+5.78$\uparrow$})
 & 72.24 (\textbf{+1.58$\uparrow$})
 & 67.83 (\textbf{+0.27$\uparrow$})
 & 72.23 (\textbf{+5.63$\uparrow$})
 & \textbf{68.75} (\textbf{+2.23$\uparrow$}) \\
\hline
\textbf{$k_2=7$, bge-m3} & \textbf{R\&B} & \textbf{R\&L} & \textbf{Full\_Ext} & \textbf{Fil} & \textbf{Full\_E\&F} & \textbf{R\&B\_Ext} & \textbf{R\&B\_Ext\_Fil} & \textbf{Average} \\
\hline
LongRAG & 59.36 & 65.56 & 68.27 & 55.31 & 67.36 & 64.42 & 63.88 & 64.45 \\
\textbf{MacRAG}
 & 62.32 (\textbf{+2.96$\uparrow$})
 & 66.34 (\textbf{+0.78$\uparrow$})
 & 71.63 (\textbf{+3.36$\uparrow$})
 & 69.72 (\textbf{+14.41$\uparrow$})
 & 73.98 (\textbf{+6.62$\uparrow$})
 & 66.61 (\textbf{+2.19$\uparrow$})
 & 72.63 (\textbf{+8.75$\uparrow$})
 & \textbf{69.03} (\textbf{+5.58$\uparrow$}) \\
\hline

\textbf{$k_2=12$, bge-m3} & \textbf{R\&B} & \textbf{R\&L} & \textbf{Full\_Ext} & \textbf{Fil} & \textbf{Full\_E\&F} & \textbf{R\&B\_Ext} & \textbf{R\&B\_Ext\_Fil} & \textbf{Average} \\
\hline
LongRAG & 60.08 & 66.77 & 69.28 & 58.50 & 69.39 & 64.90 & 65.28 & 64.89 \\
\textbf{MacRAG}
 & 64.29 (\textbf{+4.21$\uparrow$})
 & 67.20 (\textbf{+0.43$\uparrow$})
 & 70.95 (\textbf{+1.67$\uparrow$})
 & 69.46 (\textbf{+10.96$\uparrow$})
 & 73.90 (\textbf{+4.51$\uparrow$})
 & 67.49 (\textbf{+2.59$\uparrow$})
 & 71.80 (\textbf{+6.52$\uparrow$})
 & \textbf{69.30} (\textbf{+4.41$\uparrow$}) \\
\hline

\multicolumn{9}{|c|}{\textbf{Musique}} \\
\hline
\textbf{$k_2=7$, marco-miniLM} & \textbf{R\&B} & \textbf{R\&L} & \textbf{Full\_Ext} & \textbf{Fil} & \textbf{Full\_E\&F} & \textbf{R\&B\_Ext} & \textbf{R\&B\_Ext\_Fil} & \textbf{Average} \\
\hline
LongRAG & 38.98 & 41.90 & 43.64 & 37.72 & 43.83 & 42.00 & 41.97 & 41.43 \\
\textbf{MacRAG}
 & 44.76 (\textbf{+5.78$\uparrow$})
 & 45.74 (\textbf{+3.84$\uparrow$})
 & 47.42 (\textbf{+3.78$\uparrow$})
 & 47.80 (\textbf{+10.08$\uparrow$})
 & 50.09 (\textbf{+6.26$\uparrow$})
 & 48.57 (\textbf{+6.57$\uparrow$})
 & 51.00 (\textbf{+9.03$\uparrow$})
 & \textbf{47.77} (\textbf{+6.34$\uparrow$}) \\
\hline

\textbf{$k_2=12$, marco-miniLM} & \textbf{R\&B} & \textbf{R\&L} & \textbf{Full\_Ext} & \textbf{Fil} & \textbf{Full\_E\&F} & \textbf{R\&B\_Ext} & \textbf{R\&B\_Ext\_Fil} & \textbf{Average} \\
\hline
LongRAG & 41.20 & 43.85 & 48.19 & 40.48 & 47.22 & 44.66 & 44.51 & 44.30 \\
\textbf{MacRAG}
 & 45.85 (\textbf{+4.65$\uparrow$})
 & 47.66 (\textbf{+3.81$\uparrow$})
 & 48.27 (\textbf{+0.08$\uparrow$})
 & 49.01 (\textbf{+8.53$\uparrow$})
 & 48.99 (\textbf{+1.77$\uparrow$})
 & 49.35 (\textbf{+4.69$\uparrow$})
 & 50.57 (\textbf{+6.06$\uparrow$})
 & \textbf{48.53} (\textbf{+4.23$\uparrow$}) \\
\hline

\textbf{$k_2=7$, bge-m3} & \textbf{R\&B} & \textbf{R\&L} & \textbf{Full\_Ext} & \textbf{Fil} & \textbf{Full\_E\&F} & \textbf{R\&B\_Ext} & \textbf{R\&B\_Ext\_Fil} & \textbf{Average} \\
\hline
LongRAG & 42.34 & 48.08 & 47.88 & 42.17 & 47.92 & 43.70 & 44.06 & 45.16 \\
\textbf{MacRAG}
 & 45.54 (\textbf{+3.20$\uparrow$})
 & 46.68 (\textbf{-1.40$\downarrow$})
 & 49.58 (\textbf{+1.70$\uparrow$})
 & 46.76 (\textbf{+4.59$\uparrow$})
 & 49.53 (\textbf{+1.61$\uparrow$})
 & 47.94 (\textbf{+4.24$\uparrow$})
 & 49.14 (\textbf{+5.08$\uparrow$})
 & \textbf{47.88} (\textbf{+2.72$\uparrow$}) \\
\hline

\textbf{$k_2=12$, bge-m3} & \textbf{R\&B} & \textbf{R\&L} & \textbf{Ful\_Ext} & \textbf{Fil} & \textbf{Full\_E\&F} & \textbf{R\&B\_Ext} & \textbf{R\&B\_Ext\_Fil} & \textbf{Average} \\
\hline
LongRAG & 41.53 & 46.96 & 49.58 & 41.60 & 49.57 & 46.76 & 46.33 & 46.05 \\
\textbf{MacRAG}
 & 46.44 (\textbf{+4.91$\uparrow$})
 & 45.81 (\textbf{-1.15$\downarrow$})
 & 51.02 (\textbf{+1.44$\uparrow$})
 & 46.70 (\textbf{+5.10$\uparrow$})
 & 51.54 (\textbf{+1.97$\uparrow$})
 & 50.25 (\textbf{+3.49$\uparrow$})
 & 49.50 (\textbf{+3.17$\uparrow$})
 & \textbf{48.75} (\textbf{+2.70 $\uparrow$}) \\
\hline
\end{tabular}
\caption{Extensive experimental results for comparing LongRAG vs. MacRAG+LongRAG across HotpotQA, 2WikimultihopQA, and Musique datasets with two rerankers ``\textit{marco-miniLM}" and ``\textit{bge-m3}'' via GPT-4o and  F1-score. 
The experiments conducted with the same hyper-parameter ($k_1=100$, $k_2=7$) and ($k_1=100$, $k_2=12$) which is the reported best parameter of LongRAG \cite{zhao2024longrag}.
The columns represent various evaluation settings: R\&B (Retrieval and Base), R\&L (Retrieval and Long), Full\_Ext (Extraction from Full Document), Fil (Filtering), Full\_E\&F (Extraction and Filtering combined), and R\&B\_Ext (Extraction from Top-$k_2$ Chunks). Gains from applying MacRAG to LongRAG (E\&F) are displayed in parentheses with absolute gains on F1-scores and relative percentages of the improvements.}
\label{tbl_appendix_experimental_results_MacRAG_vs_LongRAG_gpt4o_detail}
\end{table*}

\end{document}